%% file: paper.tex
\pdfoutput=1
\documentclass[11pt]{article}

\usepackage{acl}

\usepackage{times}
\usepackage{latexsym}

\usepackage[T1]{fontenc}

\usepackage[utf8]{inputenc}

\usepackage{microtype}

\usepackage{xcolor}
\usepackage{xspace}
\usepackage{graphicx}
\usepackage{booktabs}
\usepackage{arydshln}
\usepackage{graphics}
\usepackage{amsmath}
\usepackage{enumitem}
\usepackage{caption}
\usepackage{subcaption}
\usepackage{multirow}
\usepackage{centernot}
\usepackage[normalem]{ulem}
\usepackage[multiple]{footmisc}
\usepackage{dblfloatfix}

\input{IJCAI_math_commands.tex} %
\newcommand{\positiveVector}{\vd^{+}}
\newcommand{\negativeVector}{\vd^{-}}
\newcommand{\queryVector}{\vd^{Q}}

\newcommand{\calL}{\mathcal{L}}

\newcommand{\dataset}{\textsc{SciDocs}\xspace}

\newcommand{\positive}{d^{+}}

\newcommand{\query}{d^{Q}}
\newcommand{\positiveCount}{c^{+}}
\newcommand{\easyNegativeCount}{c^{-}_{\text{easy}}}
\newcommand{\hardNegativeCount}{c^{-}_{\text{hard}}}
\newcommand{\positiveK}{k^{+}}

\newcommand{\hardNegativeK}{k^{-}_{\text{hard}}}

\newcommand{\map}{\textsc{map}\xspace}
\newcommand{\ndcg}{n\textsc{dcg}\xspace}

\newcommand{\knn}{\textsc{kNN}\xspace}
\newcommand{\simthreshold}{\textsc{Sim}\xspace}

\newcommand{\sys}{SciNCL\xspace} %
 
\newcommand{\oracle}{\textit{Oracle SciDocs}\xspace}

\usepackage{tikz}
\usetikzlibrary{shapes.geometric}

\definecolor{easyPosColor}{HTML}{38761d}
\definecolor{mediumPosColor}{HTML}{43b338}
\definecolor{hardPosColor}{HTML}{9de196}

\definecolor{easyNegColor}{HTML}{cc0000}
\definecolor{mediumNegColor}{HTML}{dd1010}
\definecolor{hardNegColor}{HTML}{ffdfdf} %
\definecolor{queryColor}{HTML}{2570ff}

\newbox\mybox
\def\centerfigure#1{%
    \setbox\mybox\hbox{#1}%
    \raisebox{-0.25\dimexpr\ht\mybox+\dp\mybox}{\copy\mybox}%
}
\def\centerstar#1{%
    \setbox\mybox\hbox{#1}%
    \raisebox{-0.15\dimexpr\ht\mybox+\dp\mybox}{\copy\mybox}%
}

\newbox\mybox
\def\centerfigure#1{%
    \setbox\mybox\hbox{#1}%
    \raisebox{-0.25\dimexpr\ht\mybox+\dp\mybox}{\copy\mybox}%
}
\def\centerstar#1{%
    \setbox\mybox\hbox{#1}%
    \raisebox{-0.15\dimexpr\ht\mybox+\dp\mybox}{\copy\mybox}%
}

\newcommand*\posIcon{%
       \scalebox{0.15}{
\centerfigure{\begin{tikzpicture}[]
     \draw[line width=0.75cm,black,align=center] (-1.1cm,-.5) -- (1.1cm,-0.5); %
     \draw[line width=0.75cm,black,align=center] (0,-1.6cm) -- (0, 0.6cm);
     \draw[line width=0.5cm,hardPosColor,align=center] (-1cm,-.5) -- (1cm,-0.5); %
     \draw[line width=0.5cm,hardPosColor,align=center] (0,-1.5cm) -- (0, 0.5cm);
\end{tikzpicture}}}\xspace}

\newcommand*\negIcon{%
\scalebox{0.15}{
\begin{tikzpicture}[]
    \draw[line width=0.5cm,white] (0,-.5cm) -- (0, .5cm);
    \draw[line width=0.75cm,black] (-1.1cm,0) -- (1.1cm,0); %
    \draw[line width=0.5cm,easyNegColor] (-1cm,0) -- (1cm,0);
\end{tikzpicture}}\xspace}

\newcommand*\negIconEasy{%
\scalebox{0.15}{
\begin{tikzpicture}[]
    \draw[line width=0.5cm,white] (0,-.5cm) -- (0, .5cm);
    \draw[line width=0.75cm,black] (-1.1cm,0) -- (1.1cm,0); %
    \draw[line width=0.5cm,easyNegColor] (-1cm,0) -- (1cm,0);
\end{tikzpicture}}\xspace}

\newcommand*\negIconHard{%
\scalebox{0.15}{
\begin{tikzpicture}[]
    \draw[line width=0.5cm,white] (0,-.5cm) -- (0, .5cm);
    \draw[line width=0.75cm,black] (-1.1cm,0) -- (1.1cm,0); %
    \draw[line width=0.5cm,hardNegColor] (-1cm,0) -- (1cm,0);
\end{tikzpicture}}\xspace}

\newcommand*\queryIcon{%
\scalebox{0.5}{
\centerstar{\begin{tikzpicture}[]
\node[draw=none, fill=queryColor, star, star points=5,star point ratio=2.25]  at (0,0) {};
\end{tikzpicture}}}\xspace}

\usepackage{cleveref}
\crefname{section}{\S}{\S\S}
\Crefname{section}{\S}{\S\S}
\crefname{table}{Tab.}{}
\crefname{figure}{Fig.}{}
\crefname{algorithm}{Algorithm}{}
\crefname{equation}{eq.}{}
\crefname{appendix}{Appendix}{}
\crefname{prop}{Proposition}{}
\crefformat{section}{\S#2#1#3}

\title{Neighborhood Contrastive Learning for Scientific Document Representations with Citation Embeddings}

\author{Malte Ostendorff\textsuperscript{1,2}, \hspace{1.2em} Nils Rethmeier\textsuperscript{1,3}, \hspace{1.2em} Isabelle Augenstein\textsuperscript{3}, \\
\bf{Bela Gipp\textsuperscript{2}}, \hspace{1.2em} \bf{Georg Rehm\textsuperscript{1}}
\\
\textsuperscript{1} DFKI GmbH,  
\textsuperscript{2} University of Göttingen, 
\textsuperscript{3} University of Copenhagen
\\
\textsuperscript{1} \texttt{first.lastname@dfki.de}, 
\textsuperscript{2} \texttt{lastname@uni-goettingen.de},
\\
\textsuperscript{2} \texttt{lastname@di.ku.dk}}

\begin{document}

\maketitle

\begin{abstract}

Learning scientific document representations can be substantially improved through contrastive learning objectives, where the challenge lies in creating positive and negative training samples that encode the desired similarity semantics. Prior work relies on discrete citation relations to generate contrast samples. However, discrete citations enforce a hard cut-off to similarity. This is counter-intuitive to similarity-based learning and ignores that scientific papers can be very similar despite lacking a direct citation -- a core problem of finding related research. Instead, we use controlled nearest neighbor sampling over citation graph embeddings for contrastive learning. This control allows us to learn continuous similarity, to sample hard-to-learn negatives \emph{and positives}, and also to avoid collisions between negative and positive samples by controlling the sampling margin between them. The resulting method SciNCL outperforms the state-of-the-art on the SciDocs benchmark. Furthermore, we demonstrate that it can train (or tune) language models sample-efficiently and that it can be combined with recent training-efficient methods. Perhaps surprisingly, even training a general-domain language model this way outperforms baselines pretrained in-domain.

\end{abstract}

\section{Introduction} \label{ref:intro}

\begin{figure}[ht]
\centering
\includegraphics[clip,width=0.8\linewidth]{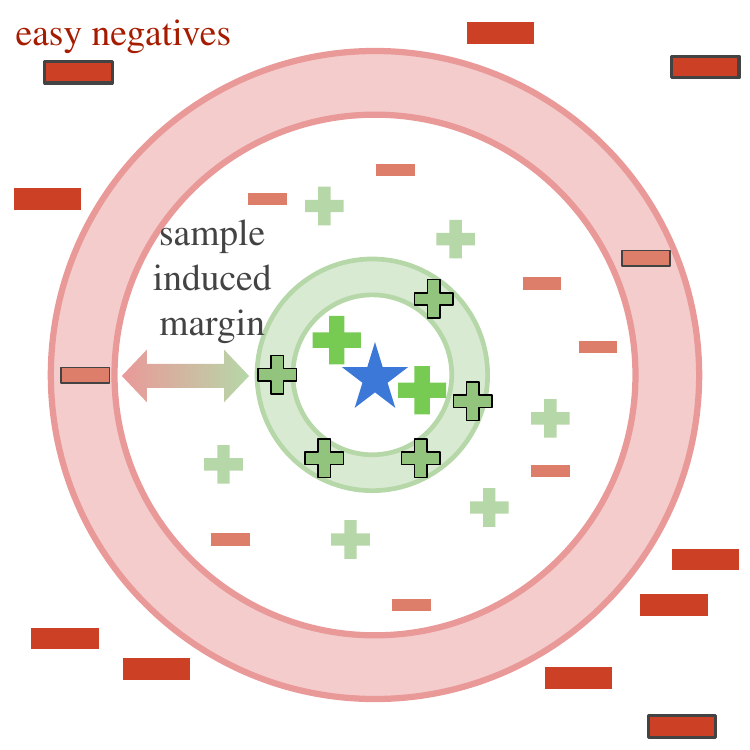}
\caption{
\label{fig:idea} Starting from a query paper \protect\queryIcon in a citation graph embedding space. 
Hard positives \protect\posIcon are citation graph embeddings that are sampled from a similar (close) context of \protect\queryIcon, but are not so close that their gradients collapse easily. 
Hard (to classify) negatives \protect\negIconHard (red band) are close to positives (green band) up to a \emph{sampling induced margin}. 
Easy negatives \protect\negIconEasy are very dissimilar (distant) from the query paper \protect\queryIcon.
}
\end{figure}

Large pretrained language models (LLMs) achieve state-of-the-art results through fine-tuning on many NLP tasks \cite{BERTOLOGY}.
However, the sentence or document embeddings derived from LLMs are of lesser quality compared to simple baselines like GloVe \cite{Reimers2019}, %
as their embedding space suffers from being anisotropic, i.e.\ poorly defined in some areas \cite{Li2020}. 

One approach that has recently gained attention is the combination of LLMs with contrastive fine-tuning to improve the semantic textual similarity between document representations \cite{Wu2020,Gao2021}. These contrastive methods learn to distinguish between pairs of similar and dissimilar texts (positive and negative samples).
As recent works show \cite{MI_views,Rethmeier2021,rethmeier2021dataefficient,TextAugemntations}, the selection of these positive and negative samples is crucial for efficient contrastive learning.

This paper focusses on learning scientific document representations (SDRs).
The core distinguishing feature of this domain is the presence of citation information that complement the textual information.
The current state-of-the-art SPECTER by \citet{Cohan2020} uses citation information to generate positive and negative samples for contrastive fine-tuning of a SciBERT language model \cite{Beltagy2019}. %
SPECTER relies on `citations by the query paper' as a discrete signal for similarity, i.e., positive samples are cited by the query while negative ones are not cited.

However, SPECTER's use of citations has its pitfalls.
Considering only one citation direction may cause positive and negative samples to collide since a paper pair could be treated as a positive and negative instance simultaneously.
Also, relying on a single citation as a discrete similarity signal is subject to noise, e.g., citations may reflect politeness and policy rather than semantic similarity \cite{Pasternack1969} or related papers lack a direct citation \cite{Gipp2009}. 
This discrete cut-off to similarity is counter-intuitive to (continuous) similarity-based learning.

Instead, the generation of non-colliding contrastive samples should be based on a continuous similarity function that allows us to find semantically similar papers, even without direct citations.
With \sys, we address these issues by generating contrastive samples based on citation embeddings.
The citation embeddings, which incorporate the full citation graph, provide a continuous, undirected, and less noisy similarity signal that allows the generations of arbitrary difficult-to-learn positive and negative samples. 

\paragraph{Contributions:}
\begin{itemize}[noitemsep]
\item%
{We propose neighborhood contrastive learning for scientific document representations with citation graph embeddings (\sys)  based on contrastive learning theory insights.} %
\item%
{We sample positive (similar) and negative (dissimilar) papers from the $k$ nearest neighbors in the citation graph embedding space, such that positives and negatives do not collide but are also hard to learn.}
\item{We compare against the state-of-the-art approach SPECTER \cite{Cohan2020} and other strong methods on the \dataset benchmark and find that \sys outperforms SPECTER on average and on 9 of 12 metrics.} 
\item{Finally, we demonstrate that with \sys, using only 1\% of the triplets for training, starting with a general-domain language model, or training only the bias terms of the model is sufficient to outperform the baselines.}
\item{Our code and models are publicly available.\footnote{\label{fn:github} \url{https://github.com/malteos/scincl}}} %
\end{itemize}

\section{Related Work} \label{sec:related-work}

\paragraph{Contrastive Learning}\ pulls representations of similar data points (positives) closer together, while representations of dissimilar documents (negatives) are pushed apart.
A common contrastive objective is the triplet loss \cite{TripletLossORG} that \citet{Cohan2020} used for scientific document representation learning, as we describe below. 
However, as \citet{MetricLearningReality} and \citet{Rethmeier2021} point out, contrastive objectives work best when specific requirements are respected. 
\textbf{(Req. 1)} Views of the same data should introduce new information, i.e.\ the mutual information between views should be minimized \cite{MI_views}. We use citation graph embeddings to generate contrast label information that supplements text-based similarity. \textbf{(Req. 2)} For training time and sample efficiency, negative samples should be hard to classify, but should also not collide with positives \cite{ContrastiveLearningLimitations}.   
\textbf{(Req. 3)} Recent works like \citet{MetricLearningReality} and \citet{muliplePositiveCTL1} use multiple positives. However, positives need to be consistently close to each other \cite{PositivesSimilarImportant}, since positives and negatives may otherwise collide, e.g., \citet{Cohan2020} consider only `citations by the query' as similarity signal and not `citations to the query'.
Such unidirectional similarity does not guarantee that a negative paper (not cited by the query) may cite the query paper and thus could cause collisions, the more we sample (Appendix~\ref{sssec:bi-specter}).
Our method treats both citing and being cited as positives (Req. 2), while it also generates hard negatives and hard positives (Req. 2+3). Hard negatives are close to but do not overlap positives (red band in \cref{fig:idea}). Hard positives are close, but not trivially close to the query document (green band in \cref{fig:idea}).
The sample induced margin (space between red and green band in \cref{fig:idea}) ensures that contrast samples do not collide.

\paragraph{Triplet Mining} remains a challenge in NLP due to the discrete nature of language which makes data augmentation less trivial as compared to computer vision \cite{Gao2021}.
Examples for augmentation strategies are translation, word deletion, or word reordering \cite{Fang2020,Wu2020}.
Positives and negatives can be sampled based on the sentence position within a document \cite{Giorgi2021}.
\citet{Gao2021} utilize supervised entailment datasets for the triplet generation.
Language- and text-independent approaches are also applied.
\citet{Kim2021} use intermediate BERT hidden state for positive sampling and \citet{Wu2021} add noise to representations to obtain negative samples.
\citet{Xiong2020} present an approach similar to \sys where they sample hard negatives from the k nearest neighbors in the embedding space derived from the previous model checkpoint.
While \citeauthor{Xiong2020} rely only on textual data, 
\sys integrates also citation information which are especially valuable in the scientific context as \citet{Cohan2020} have shown.

\paragraph{Scientific Document Representations}
based on Transformers~\cite{Vaswani2017} and pretrained on domain-specific text dominate today's scientific document processing.
There are SciBERT \cite{Beltagy2019}, BioBERT \cite{Lee2019} and SciGPT2 \cite{Luu2021}, to name a few. %
Recent works modify these domain LLMs to support cite-worthiness detection \cite{wright2021citeworth}, document similarity \cite{Ostendorff2020c}
or fact checking \cite{Wadden2020}.

Aside from text, citations are a valuable signal for the similarity of research papers.
Paper (node) representations can be learned using the citation graph \cite{Wu2019,Perozzi2014,Grover2016}.
Especially for recommendations of papers or citations, hybrid combinations of text and citation features are often employed \cite{Han2018,Jeong2020ACC,Brochier2019,Yang2015,holm2022longitudinal}.

Closest to \sys are Citeomatic \cite{Bhagavatula2018} and SPECTER \cite{Cohan2020}.
While Citeomatic relies on bag-of-words for its textual features, SPECTER is based on SciBERT.
Both leverage citations to learn a triplet-based document embedding model, whereby positive samples are papers cited in the query.
Easy negatives are random papers not cited by the query.
Hard negatives are citations of citations -- papers referenced in positive citations of the query, but are not cited directly by it.
Citeomatic also uses a second type of hard negatives, which are the nearest neighbors of a query  that are not cited by it.

Unlike our approach, Citeomatic does not use the neighborhood of citation embeddings, but instead relies on the actual document embeddings from the previous epoch.
Despite being related to \sys, the sampling approaches employed in Citeomatic and SPECTER do not account for the pitfalls of using discrete citations as signal for paper similarity.
Our work addresses this issue.

\paragraph{Cross-Modal Transfer.} 
\sys transfers knowledge across modalities, i.e., from citations into a language model.
According to \citet{Cohan2020}, \sys can be considered as a ``\textit{citation-informed Transformer}''.
This cross-modal transfer learning is applied for various modalities (see \citet{Kaur2021} for an overview):
text-to-image \cite{Socher2013},
RGB-to-depth image \cite{Tian2020},
or graph-to-image \cite{Wang2018ZeroShotRV}.
While the aforementioned methods incorporate cross-modal knowledge through joint loss functions or latent representations,
\sys transfers knowledge through the contrastive sample selection, which we found  superior to the direct transfer approach  (Appendix~\ref{sssec:student-teacher}).

\section{Methodology} \label{sec:methodology}
Our goal is to learn citation-informed representations for scientific documents. To do so we sample three document representation vectors and learn their similarity. For a given query paper vector $\queryVector$, we sample a positive (similar) paper vector $\positiveVector$ and a negative (dissimilar) paper vector $\negativeVector$. This produces a `query, positive, negative' triplet $(\queryVector,\positiveVector,\negativeVector)$ -- represented by (\queryIcon,\posIcon,\negIcon) in \cref{fig:idea}. To learn paper similarity, we need to define three components: (\cref{ssec:cl-objective}) how to calculate document vectors $\vd$ for the loss over triplets $\calL$; (\cref{ssec:CGEmbeddings}) how citations provide similarity between papers; and (\cref{ssec:neg_pos_sampling}) how negative and positive papers $(\negativeVector,\positiveVector)$ are sampled as (dis-)similar documents from the neighborhood of a query paper $\queryVector$.

\subsection{Contrastive Learning Objective} \label{ssec:cl-objective} %
Given the textual content of a document $d$ (paper), the goal is to derive a dense vector representation $\vd$ that best encodes the document information and can be used in downstream tasks.
A Transformer language model $f$ (SciBERT; \citet{Beltagy2019}) encodes documents $d$ into vector representations $f(d)=\vd$.
The input to the language model is the title and abstract separated by the \texttt{[SEP]} token.\footnote{\citet{Cohan2019} evaluated other inputs (venue or author) but found the title and abstract to perform best.}
The final layer hidden state of the \texttt{[CLS]} token is then used as a document representation $f(d)=\vd$.

Training with a masked language modeling objectives alone has been shown to produce sub-optimal document representations \cite{Li2020,Gao2021}. Thus, similar to the SDR state-of-the-art method SPECTER \cite{Cohan2020}, we continue training the SciBERT model \cite{Beltagy2019} using a self-supervised triplet margin loss \cite{TripletLossORG}:
\begin{equation*}
    \calL=
    \max\Big\{
        \lVert \queryVector{-}\positiveVector \lVert_2{-}\lVert \queryVector{-}\negativeVector \lVert_2
        + \xi,0\Big\}
\label{eq:loss}
\end{equation*}
\noindent Here, $\xi$ is a slack term ($\xi=1$ as in SPECTER) and $\lVert\Delta\vd\lVert_2$ is the $L^2$ norm, used as a distance function. However, the SPECTER sampling method has significant drawbacks. We will describe these issues and our contrastive learning theory guided improvements in detail below in \cref{ssec:CGEmbeddings}.

\subsection{Citation Neighborhood Sampling} \label{ssec:CGEmbeddings}

Compared to the textual content of a paper, citations provide an outside view on a paper and its relation to the scientific literature  \cite{Elkiss2008}, which is why citations are traditionally used as a similarity measure in library science \cite{Kessler1963,Small1973}.
However, using citations as a discrete similarity signal, as done in \citet{Cohan2020}, has its pitfalls. Their method defines papers cited by the query as positives, while paper citing the query could be treated as negatives. This means that \emph{positive and negative learning information collides} between citation directions, which \citet{ContrastiveLearningLimitations} have shown to deteriorate performance.  
Furthermore, a cited paper can have a low similarity with the citing paper given the many motivations a citation can have \cite{Teufel2006}.
Likewise, a similar paper might not be cited. 

To overcome these limitations, we learn citation embeddings first and then use the citation neighborhood around a given query paper $d^Q$ to construct similar (positive) and dissimilar (negative) samples for contrast 
by using the $k$ nearest neighbors.
This builds on the intuition that nodes connected by edges should be close to each other in the embedding space \cite{Perozzi2014}. %
Using citation embeddings allows us to: (1) sample paper similarity on a continuous scale, which makes it possible to: (2) define hard to learn positives, as well as (3) hard or easy to learn negatives. Points (2-3) are important in making contrastive learning efficient as will describe below in \cref{ssec:neg_pos_sampling}.

\subsection{Positives and Negatives Sampling}\label{ssec:neg_pos_sampling}
\paragraph{Positive samples:} $\positive$ should be semantically similar to the query paper $\query$, i.e.\ sampled close to the query embedding $\queryVector$. Additionally, as \citet{PositivesSimilarImportant} find, positives should be sampled from comparable locations (distances from the query) in embedding space and be dissimilar enough from the query embedding, to avoid gradient collapse (zero gradients). 
Therefore, we sample $\positiveCount$ positive (similar) papers from a close neighborhood around query embedding $\queryVector$ $(\positiveK-\positiveCount,\positiveK]$,
i.e.\ the green band in \cref{fig:idea}. When sampling with \knn search, we use a small $\positiveK$ to find positives and later analyze the impact of $\positiveK$ in \cref{fig:easy_pos}.

\paragraph{Negative samples:}
can be divided into easy \negIcon and hard \negIconHard negative samples (light and dark red in \cref{fig:idea}). Sampling more hard negatives is known to improve contrastive learning \cite{Bucher2016,Wu2017}.
However, we make sure to sample hard negatives (red band in \cref{fig:idea}) such that they are close to potential positives but do not collide with positives (green band), by using a tunable `sampling induced margin'.
We do so, since \citet{ContrastiveLearningLimitations} showed that sampling a larger number of hard negatives only improves performance \emph{if the negatives do not collide with positive samples}, since collisions make the learning signal noisy.
That is, in the margin between hard negatives and positives we expect positives and negatives to collide, thus we avoid sampling from this region. To generate a diverse self-supervised citation similarity signal for contrastive SDR learning, we also sample easy negatives that are farther from the query than hard negatives.
For negatives, the $k^{-}$ should be large when sampling via \knn
to ensure samples are dissimilar from the query paper.

\subsection{Sampling Strategies} \label{sssec:sampling-strategies} \label{ssec:triplet-minig}

As described in \cref{ssec:CGEmbeddings} and \cref{ssec:neg_pos_sampling}, our approach improves upon the method by \citet{Cohan2020}. Therefore, we reuse their sampling parameters (5 triplets per query paper) and then further optimize our methods' hyperparameters.
For example, to train the triplet loss, we generate the same amount of $(\queryVector, \positiveVector, \negativeVector)$ triplets per query paper as SPECTER \cite{Cohan2020}. To be precise, this means we generate $\positiveCount{=}5$ positives (as explained in \cref{ssec:neg_pos_sampling}). We also generate 5 negatives, three easy negatives $\easyNegativeCount{=}3$ and two hard negatives $\hardNegativeCount{=}2$, as described in \cref{ssec:neg_pos_sampling}.

Below, we describe three strategies (I-III) for sampling triplets. 
These either sample neighboring papers from citation embeddings (I), by random sampling (II), or using both strategies (III).
For each strategy, let $c'$ be the number of samples for either positives $\positiveCount$, easy negatives $\easyNegativeCount$, or hard negatives $\hardNegativeCount$.

\paragraph{Citation Graph Embeddings:}
We train a graph embedding model $f_c$ on citations extracted from the Semantic Scholar Open Research Corpus \citep[S2ORC; ][]{Lo2020} to get citation embeddings $C$.
We utilize PyTorch BigGraph \cite{Lerer2019}, which allows for training on large graphs with modest hardware requirements.
The resulting graph embeddings perform well using the default training settings from \citet{Lerer2019}, but given more computational resources, careful tuning may produce even better-performing embeddings.
Nonetheless, we conducted a narrow parameter search based on link prediction -- see Appendix~\ref{ssec:graph-evaluation}.

\paragraph{(I) K-nearest neighbors (\knn):} 
Assuming a given citation embedding model $f_c$ and a search index (e.g., FAISS \cref{ssec:Train_implement}), we run $KNN(f_c(\query), C)$ and take $c'$ samples from a range of the $(k-c',k]$ nearest neighbors around the query paper
$\query$ with its neighbors $N{=}\{n_1,n_2,n_3,\dots\}$, whereby neighbor $n_i$ is the $i$-th nearest neighbor in the citation embedding space.
For instance, for $c'{=}3$ and $k{=}10$ the corresponding samples would be the three neighbors descending from the tenth neighbor: $n_8$, $n_9$, and $n_{10}$. 
To reduce computing effort, we sample the neighbors $N$ only once via $[0;\max(\positiveK,\hardNegativeK)]$, and then generate triplets by range-selection in $N$; i.e.\ positives $=(\positiveK-\positiveCount;\positiveK]$, and hard negatives $=(\hardNegativeK-\hardNegativeCount;\hardNegativeK]$.

\paragraph{(II) Random sampling:} 
Sample any $c'$ papers without replacement from the corpus. 

\paragraph{(III) Filtered random:} 
Like (II) but excluding the papers that are retrieved by \knn, i.e., all neighbors within the largest $k$ are excluded.
This is analog to SPECTER's approach of selecting random candidates that are not cited by the query. 

\vspace{0.25cm}
The \knn sampling introduces the hyperparameter $k$ that allows for the \emph{controlled sampling of positives or negatives} with different difficulty (from easy to hard depending on $k$).
Specifically, in \cref{fig:idea} the hyperparameter $k$ defines the tunable \emph{sample induced margin} between positives and negatives, as well as the width and position of the positive sample band (green) and negative sample band (red) around the query sample.
Besides the strategies above, we experiment with similarity threshold, k-means clustering and sorted random sampling, neither of which performs well (Appendix~\ref{ssec:negative-results}).

\section{Experiments} \label{sec:experiments}

In the following, we introduce our experiments including the data sets and implementation details.

\subsection{Evaluation Dataset}

We evaluate on the \dataset benchmark \cite{Cohan2020}.
A key difference to other benchmarks is that embeddings are the input to the individual tasks without explicit fine-tuning.
The \dataset benchmark consists of the following four tasks:

\textbf{Document classification} (CLS) with Medical Subject Headings (MeSH) \cite{Lipscomb2000MedicalSH} and Microsoft Academic Graph labels (MAG) \cite{Sinha2015AnOO}. %
\textbf{Co-views and co-reads} (USR) prediction based on the L2 distance between embeddings. 
\textbf{Direct and co-citation} (CITE) prediction based on the L2 distance between the embeddings.
\textbf{Recommendations} (REC) generation based on embeddings and paper metadata.

\subsection{Training Datasets} \label{ssec:training-data}

The experiments mainly compare \sys against SPECTER on the \dataset benchmark. 
However, we found 40.5\% of \dataset's papers leaking into SPECTER's training data (the leakage affects only the unsupervised paper data but not the gold labels -- see Appendix~\ref{ssec:leakage}).
To be transparent about this leakage, we train \sys on two datasets:

\paragraph{SPECTER replication (w/ leakage):}
We replicate SPECTER's training data including its leakage.
Unfortunately, SPECTER provides neither citation data nor a mapping to S2ORC, which our citation embeddings are based on.
We successfully map 96.2\% of SPECTER's query papers and 83.3\% of the corpus from which positives and negatives are sampled to S2ORC.
To account for the missing papers, we randomly sample papers from S2ORC (without the \dataset papers) such that the absolute number of papers is identical with SPECTER.

\paragraph{S2ORC subset (w/o leakage):}
We select a random subset from S2ORC that does not contain any of the mapped \dataset papers.
This avoids SPECTER's leakage, but also makes the scores reported in \citet{Cohan2020} less comparable.
We successfully map 98.6\% of the \dataset papers to S2ORC. 
Thus, only the remaining 1.4\% of the \dataset papers could leak into this training set.

The details of the dataset creation are described in Appendix~\ref{s2orc-mapping} and \ref{ssec:dataset-creation}.
Both training sets yield 684K triplets (same count as SPECTER).
Also, the ratio of training triplets per query remains the same (\cref{sssec:sampling-strategies}).
Our citation embedding model is trained on the S2ORC citation graph.
In \textit{w/ leakage}, we include all SPECTER papers even if they are part of \dataset, the remaining \dataset papers are excluded (52.5 nodes and 463M edges).
In \textit{w/o leakage}, all mapped \dataset papers are excluded (52.4M nodes and 447M edges) such that we avoid leakage also for the citation embedding model.

\subsection{Model Training and Implementation}\label{ssec:Train_implement}

We replicate the training setup from SPECTER as closely as possible.
We implement \sys using Huggingface Transformers \cite{Wolf2019}, initialize the model with SciBERT's weights \cite{Beltagy2019}, and train via the triplet loss (Equation~\ref{eq:loss}). 
The optimizer is Adam with weight decay \cite{Kingma2015,Loshchilov2019} and learning rate $\lambda{=}2^{-5}$.
To explore the effect of computing efficient fine-tuning we also train a BitFit model \cite{Zaken2021} with $\lambda{=}1^{-4}$ (\cref{ssec:data-efficiency}).
We train \sys on two NVIDIA GeForce RTX 6000 (24G) for 2 epochs (approx. 24 hours of training time) with batch size 8 and gradient accumulation for an effective batch size of 32 (same as SPECTER). 
The graph embedding training is performed on an Intel Xeon Gold 6230 CPU with 60 cores and takes approx. 6 hours.
The \knn strategy is implemented with FAISS \cite{Johnson2021} using a flat index (exhaustive search) and takes less than 30min for indexing and retrieval of the triplets.

\subsection{Baseline Methods} \label{ssec:baselines}

\begin{table*}[ht!]
\centering
\footnotesize
\setlength{\tabcolsep}{4.7pt}
\renewcommand{\arraystretch}{1.3}
\begin{tabular}{@{}lrrrrrrrrrrrrr@{}}
\toprule
Task $\rightarrow$                 & \multicolumn{2}{c}{Classification} & \multicolumn{4}{c}{User activity prediction}                        & \multicolumn{4}{c}{Citation prediction}                & \multicolumn{2}{c}{\multirow{2}{*}{\begin{tabular}{@{}c@{}}Recomm.\end{tabular}}} & 
\multirow{3}{*}{\begin{tabular}{@{}c@{}}Avg.\end{tabular}}
\\ \cmidrule(lr){2-3} \cmidrule(lr){4-7} \cmidrule(lr){8-11} 
Subtask $\rightarrow$              & MAG              & MeSH            & \multicolumn{2}{c}{Co-View} & \multicolumn{2}{c}{Co-Read} & \multicolumn{2}{c}{Cite} & \multicolumn{2}{c}{Co-Cite} & \multicolumn{2}{c}{} &                   \\ \cdashline{1-13}
Model $\downarrow$ / Metric $\rightarrow$               & F1               & F1              & \map        & \ndcg        & \map         & \ndcg        & \map       & \ndcg       & \map         & \ndcg        & \ndcg        & P@1       &  \\ 
\midrule

\oracle $\dagger$ 
 & \textit{87.1}  & \textit{94.8}  & \textit{87.2}  & \textit{93.5}  & \textit{88.7}  & \textit{94.6}  & \textit{92.3}  & \textit{96.8}  & \textit{91.4}  & \textit{96.4}  & \textit{53.8}  & \textit{19.4}  & \textit{83.0} \\
 
 USE \citeyearpar{Cer2018UniversalSE} &               80.0 &                83.9 &                    77.2 &                     88.1 &                    76.5 &                     88.1 &                 76.6 &                  89.0 &                    78.3 &                     89.8 &                        53.7 &                       19.6 &           75.1 \\
Citeomatic* \citeyearpar{Bhagavatula2018}           & 67.1 & 75.7 & 81.1 & 90.2 & 80.5 & 90.2 & 86.3 & 94.1 & 84.4 & 92.8 & 52.5 & 17.3 & 76.0 \\ 
SGC* \citeyearpar{Wu2019}               & 76.8 & 82.7 & 77.2 & 88.0 & 75.7 & 87.5 & 91.6 & 96.2 & 84.1 & 92.5 & 52.7 & 18.2 & 76.9 \\
BERT \citeyearpar{Devlin2019}  &               79.9 &                74.3 &                    59.9 &                     78.3 &                    57.1 &                     76.4 &                 54.3 &                  75.1 &                    57.9 &                     77.3 &                        52.1 &                       18.1 &           63.4 \\
SciBERT* \citeyearpar{Beltagy2019} & 79.7 & 80.7 & 50.7 & 73.1 & 47.7 & 71.1 & 48.3 & 71.7 & 49.7 & 72.6 & 52.1 & 17.9 & 59.6 \\ 
BioBERT \citeyearpar{Lee2019} &               77.2 &                73.0 &                    53.3 &                     74.0 &                    50.6 &                     72.2 &                 45.5 &                  69.0 &                    49.4 &                     71.8 &                        52.0 &                       17.9 &           58.8 \\
CiteBERT \citeyearpar{wright2021citeworth} &             78.8 &                74.8 &                    53.2 &                     73.6 &                    49.9 &                     71.3 &                 45.0 &                  67.9 &                    50.3 &                     72.1 &                        51.6 &                       17.0 &           58.8 \\
DeCLUTR \citeyearpar{Giorgi2021} &               81.2 &                88.0 &                    63.4 &                     80.6 &                    60.0 &                     78.6 &                 57.2 &                  77.4 &                    62.9 &                     80.9 &                        52.0 &                       17.4 &           66.6 \\

SPECTER* \citeyearpar{Cohan2020}         &  \bf{82.0} &	86.4 & 83.6 & 91.5 & 84.5 & 92.4 & 88.3 &	94.9 & 88.1 & 94.8 & \textbf{53.9}& \bf{20.0} &  80.0 \\

\hdashline

\multicolumn{14}{@{}p{0.8\textwidth}}{\textit{Replicated SPECTER training data (w/ leakage):}} \\

\sys (ours) & 81.4& \textbf{88.7} & \textbf{85.3} & \textbf{92.3} & \textbf{87.5} & \textbf{93.9} & \textbf{93.6} & \textbf{97.3} & \textbf{91.6} & \textbf{96.4} & \textbf{53.9} & 19.3& \textbf{81.8}  \\ 
$\pm\ \sigma$ w/ ten seeds& \textit{.449} & \textit{.422} & \textit{.128} & \textit{.08} & \textit{.162} & \textit{.118} & \textit{.104} & \textit{.054} & \textit{.099} & \textit{.066} & \textit{.203} & \textit{.356} & \textit{.064}  \\ 
\hdashline
\multicolumn{14}{@{}p{0.8\textwidth}}{\textit{Random S2ORC training data (w/o leakage):}} \\

SPECTER & 81.3  & 88.4  & 83.1  & 91.3  & 84.0  & 92.1  & 86.2  & 93.9  & 87.8  & 94.7  & 52.2  & 17.5  & 79.4 \\
\sys (ours)  & 81.3  & 89.4  & 84.3  & 91.8  & 85.6  & 92.8  & 91.4  & 96.3  & 90.1  & 95.7  & 54.3  & 19.9  & 81.1  \\

\bottomrule
\end{tabular}
\caption{Results on the \dataset test set. 
With replicated SPECTER training data, \sys surpasses the previous best avg. score by 1.8 points and also outperforms the baselines in 9 of 12 task metrics. 
Our scores are reported as mean and standard deviation $\sigma$ over ten random seeds. 
With training data randomly sampled from S2ORC, \sys outperforms SPECTER in terms of avg. score with 1.7 points.
The scores with * are from \citet{Cohan2020}.
\oracle $\dagger$ is the upper bound of the performance with triplets from \dataset's data.}
\label{tab:results}
\end{table*}

We compare against the following baselines (details in Appendix \ref{ssec:baseline-details}):
USE~\cite{Cer2018UniversalSE},
BERT \cite{Devlin2019}, %
BioBERT \cite{Lee2019}, %
SciBERT \cite{Beltagy2019}, %
CiteBERT \cite{wright2021citeworth}, %
DeCLUTR \cite{Giorgi2021},
the graph-convolution approach SGC \cite{Wu2019}, 
Citeomatic \cite{Bhagavatula2018}, and SPECTER \cite{Cohan2020}.

Also, we compare against \oracle which is identical to \sys except that its triplets are generated based on \dataset's validation and test set using their gold labels.
For example, papers with the same MAG labels are positives and papers with different labels are negatives.
Similarly, the ground truth of the other tasks is used, i.e., clicked recommendations are considered as positives etc.
In total, this procedure creates 106K training triplets for \oracle. 
Moreover, we under-sample triplets from the classification tasks to ensure a balanced triplet distribution over the tasks.
Accordingly, \oracle represents an estimate for the performance upper bound that can be achieved with the current setting (triplet margin loss and SciBERT encoder).

\section{Overall Results}

\cref{tab:results} shows the results, comparing \sys with the best validation performance against the baselines.
With replicated SPECTER training data (w/ leakage), \sys achieves an average performance of 81.8 across all metrics, which is a 1.8 point absolute improvement over SPECTER (the next-best baseline). 
When trained without leakage, the improvement of \sys over SPECTER is consistent with 1.7 points but generally lower (79.4 avg. score).
In the following, we refer to the results obtained through training on the replicated SPECTER data (w/ leakage) if not otherwise mentioned.

We find the best validation performance based on SPECTER's data when positives and hard negative are sampled with \knn, whereby positives are $\positiveK{=}25$, and hard negatives are $\hardNegativeK{=}4000$ (\cref{sec:hyperparameter-analysis}).
Easy negatives are generated through filtered random sampling.
\sys's scores are reported as mean over ten random seeds ($\text{seed}\in[0,9]$).

For MAG classification, SPECTER achieves the best result with 82.0 F1 followed by \sys with 81.4 F1 (-0.6 points).
For MeSH classification, \sys yields the highest score with 88.7 F1 (+2.3 compared to SPECTER).
Both classification tasks have in common that the chosen training settings lead to over-fitting.
Changing the training by using only 1\% training data, \sys yields 82.2 F1@MAG (\cref{tab:ablation}). %
In all user activity and citation tasks, \sys yields higher scores than all baselines.
Moreover, \sys outperforms SGC on direct citation prediction, where SGC outperforms SPECTER in terms of \ndcg.
On the recommender task, SPECTER yields the best P@1 with 20.0, whereas \sys achieves 19.3 P@1 (in terms of \ndcg \sys and SPECTER are on par).

When training SPECTER and \sys without leakage, \sys outperforms SPECTER even in 11 of 12 metrics and is on par in the other metric.
This suggests that \sys's hyperparameters have a low corpus dependency since they were only optimized on the corpus with leakage.

Regarding the LLM baselines, we observe that the general-domain BERT, with a score of 63.4, outperforms the domain-specific BERT variants, namely SciBERT (59.6) and BioBERT (58.8). %
LLMs without citations or contrastive objectives yield generally poor results.
This emphasizes the anisotropy problem of embeddings directly extracted from current LLMs and highlights the advantage of combining text and citation information. %

In summary, we show that \sys's triplet selection leads on average to a performance improvement on \dataset, with most gains being observed for user activity and citation tasks.
The gain from 80.0 to 81.8 is particularly notable  given that even \oracle yields with 83.0 an only marginally higher avg. score despite using test and validation data from \dataset for the triplet selection.
Appendix~\ref{ssec:examples} shows examples of paper triplets.

\section{Impact of Sample Difficulty} \label{sec:hyperparameter-analysis}

In this section, we present the optimization of \sys's sampling strategy (\cref{ssec:neg_pos_sampling}).
We optimize the sampling for positives and hard or easy negatives with partial grid search on a random sample of 10\% of the replicated SPECTER training data (sampling based on queries).
Our experiments show that optimizations on this subset correlate with the entire dataset.
The validation scores in \cref{fig:easy_pos} and \ref{fig:hard_neg} are reported as the mean over three random seeds.

\subsection{Positive Samples}  \label{ssec:positves-results}

\begin{figure}[h]
\centering
\includegraphics[clip,width=1.\linewidth]{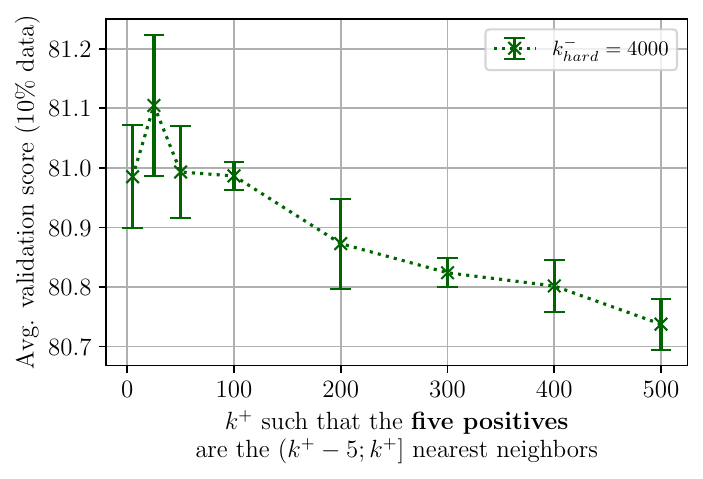}
\caption{\label{fig:easy_pos}Results on the validation set  w.r.t. positive sampling with \knn when using 10\% training data.}
\end{figure}

\cref{fig:easy_pos} shows the avg. scores on the \dataset validation set depending on the selection of positives with the \knn strategy. 
We only change $\positiveK$, while negative sampling remains fixed to its best setting (\cref{ssec:hard-negatives-results}).
The performance is relatively stable for $\positiveK{<}100$ with peak at $\positiveK{=}25$, for $\positiveK{>}100$ the performance declines as $\positiveK$ increases.
\citet{PositivesSimilarImportant} state that positive samples should be semantically similar to each other, but not too similar to the query. For example, at $\positiveK{=}5$, positives may be a bit ``too easy'' to learn, such that they produce less informative gradients than the optimal setting $\positiveK{=}25$.
Similarly, making $\positiveK$ too large leads to the \emph{sampling induced margin} being too small, such that \emph{positives collide with negative samples}, which creates contrastive label noise that degrades performance \citep{ContrastiveLearningLimitations}. 

Another observation is the standard deviation $\sigma$: %
One would expect $\sigma$ to be independent of $\positiveK$ since random seeds affect only the negatives.
However, positives and negatives interact with each other through the triplet margin loss.
Therefore, $\sigma$ is also affected by $\positiveK$. 
To account for the interaction of positives and negatives, one could sample simultaneously based on the distance to the query and the distance of positives and negatives to each other.

\subsection{Hard Negative Samples} \label{ssec:hard-negatives-results}

\begin{figure}[h]
\centering
\includegraphics[clip,width=1.\linewidth]{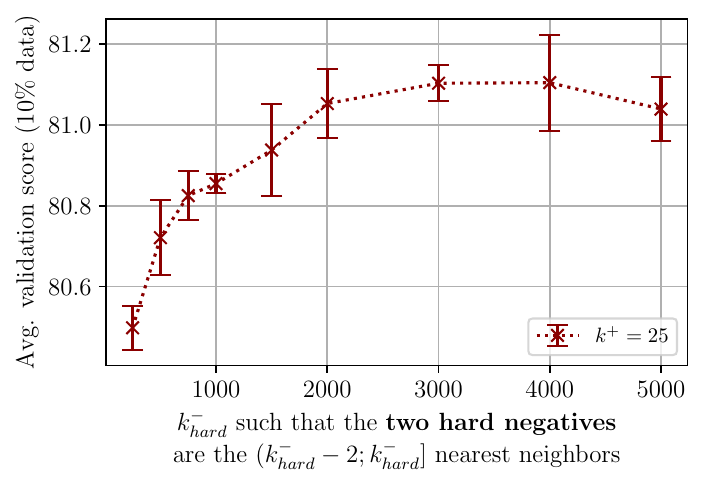}
\caption{\label{fig:hard_neg}Results on the validation set w.r.t. hard negative sampling with \knn using 10\% training data.}
\end{figure}

\cref{fig:hard_neg} presents the validation results for different $\hardNegativeK$ given the best setting for positives ($\positiveK{=}25$).
The performance increases with increasing $\hardNegativeK$ until a plateau between $2000{<}\hardNegativeK{<}4000$ with a peak at $\hardNegativeK{=}4000$. 
This plateau can also be observed in the test set, where $\hardNegativeK{=}3000$ yields a marginally lower score of 81.7 (\cref{tab:ablation}).
For $\hardNegativeK{>}4000$, the performance starts to decline again.
This suggests that for large $\hardNegativeK$ the samples are not ``hard enough'' which confirms the findings of \citet{Cohan2020}.

\subsection{Easy Negative Samples}

Filtered random sampling of easy negatives yields the best validation performance compared pure random sampling (\cref{tab:ablation}).
However, the performance difference is marginal.
When rounded to one decimal, their average test scores are identical.
The marginal difference is caused by the large corpus size and the resulting small probability of randomly sampling one paper from the \knn results.
But without filtering, the effect of random seeds increases, since we find a higher standard deviation compared to the one with filtering. %

As a potential way to decrease randomness, we experiment with other approaches like k-means clustering but find that they decrease the performance (Appendix~\ref{ssec:negative-results}).

\subsection{Collisions} \label{ssec:collisions}

\begin{figure}[ht]
\centering
\includegraphics[clip,width=1.\linewidth]{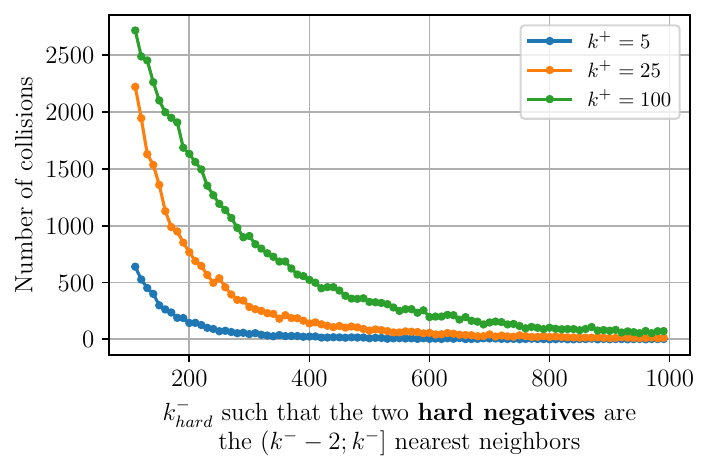}

\caption{\label{fig:collisions}Number of collisions w.r.t. size of the sample induced margin as defined through  $\positiveK$ and $\hardNegativeK$.}
\end{figure}

Similar to SPECTER, SciNCL's sampling based on graph embeddings could cause collisions when selecting positives and negatives from regions close to each other. 
To avoid this, we rely on a sample induced margin that is defined by the hyperparameter $\positiveK$ and $\hardNegativeK$ (distance between red and green band in \cref{fig:idea}).
When the margin gets too small, positives and negatives are more likely to collide. 
A collision occurs when the paper pair $(d_q,d_s)$ is contained in the training data as positive and as negative sample at the same time.
\cref{fig:collisions} demonstrates the relation between the number of collisions and the size of the sample induced margin. 
The number of collisions increases when the sample induced margin gets smaller.
The opposite is the case  when the margin is large enough ($\hardNegativeK>1000$), i.e., then the number of collisions goes to zero.
This relation also affects the evaluation performance as \cref{fig:easy_pos} and \cref{fig:hard_neg} show. 
Namely, for large $\positiveK$ or small  $\hardNegativeK$ SciNCL's performance declines and approaches SPECTER's performance.

\section{Ablation Analysis} \label{sec:ablation-analysis}

Next, we evaluate the impact of language model initialization and number of parameters and triples.

\begin{table}[!t]
\footnotesize
\centering
\setlength{\tabcolsep}{3pt}
\renewcommand{\arraystretch}{1.2}
\begin{tabular}{@{}lrrrr|rr@{}}
\toprule 
 &CLS       & USR   & CITE  & REC  & Avg. & $\Delta$   \\ \midrule

\sys & 85.0  & 88.8  & \textbf{94.7}  & 36.6  & \textbf{81.8}  & -- \\ 
SPECTER & 84.2 & 88.4 & 91.5 & 36.9 & 80.0 & -1.8\\ 
 \hdashline
 
$\hardNegativeK{=}2000$ & 84.9  & 88.8  & \textbf{94.7}  & 36.1  & 81.6  & -0.2  \\ 

$\hardNegativeK{=}3000$ & 84.5  & 88.7  & 94.6  & 36.9  & 81.7  & -0.1  \\

easy neg. w/ random & 85.1  & 88.8  & \textbf{94.7}  & 36.6  & \textbf{81.8}  & 0.0  \\ 

undirected citations & 84.6  & 88.8  & \textbf{94.7}  & 36.6  & 81.7  & -0.1  \\ 

\hdashline

Init. w/ BERT-Base  & 83.4  & 88.4  & 93.8  & 37.5  & 81.2  & -0.6  \\ 
Init. w/ BERT-Large  & 84.6  & 88.7  & 94.1  & 36.4  & 81.4  & -0.4  \\ 
Init. w/ BioBERT  & 83.7  & 88.6  & 93.8  & \textbf{37.7}  & 81.4  & -0.4  \\ 

\hdashline

1\% training data  & 85.2  & 88.3  & 92.7  & 36.1  & 80.8  & -1.0  \\ 

10\% training data & 85.1  & 88.7  & 93.5  & 36.2  & 81.1  & -0.6  \\ 

BitFit training & \textbf{85.8}  & 88.6  & 93.7  & 35.3  & 81.2  & -0.5  \\

\bottomrule
\end{tabular}
\caption{Ablations. Numbers are averages over tasks of the \dataset test set, average score over all metrics, and rounded absolute difference to \sys.}
\label{tab:ablation}
\end{table}

\subsection{Initial Language Models} \label{ssec:init-lms}

\cref{tab:ablation} shows the effect of initializing the model weights not with SciBERT but with general-domain LLMs (BERT-Base and BERT-Large) or with BioBERT. 
The initialization with other LLMs decreases the performance.
However, the decline is marginal (BERT-Base -0.6, BERT-Large -0.4, BioBERT -0.4) and all LLMs outperform the SPECTER baseline.
For the recommendation task, in which SPECTER is superior over \sys, BioBERT outperforms SPECTER. %
This indicates that the improved triplet mining of \sys has a greater domain adaption effect than pretraining on domain-specific literature.
Given that pretraining of LLMs requires a magnitude more resources than the fine-tuning with \sys, our approach can be a solution for resource-limited use cases.

\subsection{Data and Computing Efficiency} \label{ssec:data-efficiency} %

The last three rows of \cref{tab:ablation} show the results regarding data and computing efficiency.
When keeping the citation graph unchanged but training the language model with only 10\% of the original triplets, \sys still yields a score of 81.1 (-0.6).
Even with only 1\% (6840 triplets), \sys achieves a score of 80.8 that is 1.0 points less than with 100\% but still 0.8 points more than the SPECTER baseline.
With this \textit{textual} sample efficiency, one could manually create triplets or use existing supervised datasets as in \citet{Gao2021}.

Lastly, we evaluate BitFit training \cite{Zaken2021}, which only trains the bias terms of the model while freezing all other parameters.
This corresponds to training only 0.1\% of the original parameters.
With BitFit, \sys yields a considerable score of 81.2 (-0.5 points).
As a result, \sys could be trained  on the same hardware with even larger (general-domain) language models (\cref{ssec:init-lms}).

\section{Conclusion}

We present a novel approach for contrastive learning of scientific document embeddings that addresses the challenge of selecting informative positive and negative samples.
By leveraging citation graph embeddings for sample generation, \sys achieves a score of 81.8 on the \dataset benchmark, a 1.8 point improvement over the previous best method SPECTER.
This is purely achieved by introducing tunable sample difficulty and avoiding collisions between positive and negative samples, while existing LLM and data setups can be reused. 
This improvement over SPECTER can be also observed when excluding the \dataset papers during training (see w/o leakage in \cref{tab:results}). 
Furthermore, \sys's improvement from 80.0 to 81.8 is particularly notable given that even \textit{oracle triplets}, which are generated with \dataset's test and validation data, yield with 83.0 only a marginally higher score. %

Our work highlights the importance of sample generation in a contrastive learning setting.
We show that language model training with 1\% of triplets is sufficient to outperform SPECTER, whereas the remaining 99\% provide only 1.0 additional points (80.8 to 81.8).
This sample efficiency is achieved by adding reasonable effort for sample generation, i.e., graph embedding training and \knn
search. 
We also demonstrate that in-domain LLM pretraining (like SciBERT) is beneficial, while general-domain LLMs can achieve comparable performance and even outperform SPECTER.
This indicates that controlling sample difficulty and avoiding collisions
is more effective than in-domain pretraining, especially in scenarios where training an LLM from scratch is infeasible.

\section{Limitations}

\sys's strategy of selecting positive and negative samples requires additional computational resources for training the graph embedding model, performing the $\knn$ search, and optimizing the hyperparameters $\positiveK,\hardNegativeK$ (\cref{ssec:Train_implement}). %
While some of the compute resources are offset by the sample-efficient language model training (\cref{ssec:data-efficiency}), %
we  still consider the increased compute effort as the major limitation of the \sys method.

Especially the training of the graph embedding model accounts for most of the additional compute effort.
This is also the reason for us providing only a shallow of evaluation of the graph embeddings (\cref{ssec:graph-evaluation}). %
For example, we did not evaluate the effect of different graph embeddings on the actual \dataset performance.
Moreover, evaluations with smaller subsets of the S2ORC citation graph are missing.
Such evaluations could indicate whether also less citation data can be sufficient, which would lower the compute requirements but would  make \sys also applicable in domains where less graph data is available.

\section*{Acknowledgements}

We would like to thank Christian Schulze and his team for providing the compute infrastructure that made our experiments possible.
The research presented in this article is partially funded by the German Federal Ministry of Education and Research (BMBF) through the projects QURATOR \cite{rehm2020d} (Unternehmen Region, Wachstumskern, no.~03WKDA1A) and PANQURA (no.~03COV03E).

\bibliography{custom}

\appendix

\clearpage

\label{sec:appendix}

\section{Mapping to S2ORC} \label{s2orc-mapping}

\begin{table}[ht]
\centering
\footnotesize
\caption{Mapping to S2ORC citation graph}
\label{tab:id-mapping}
\begin{tabular}{lr}
\toprule
\multicolumn{1}{c}{\textbf{S2ORC mapping}} & \multicolumn{1}{c}{\textbf{Success rate}} \\
\midrule
SciDocs papers          & \multicolumn{1}{l}{}       \\
- with S2ORC IDs        & 220,815 / 223,932 (98.6\%) \\
- in S2ORC graph & 197,811 / 223,932 (88.3\%) \\
\rule{0pt}{4ex}    
SPECTER papers          & \multicolumn{1}{l}{}       \\
- with S2ORC IDs        & 311,094 / 311,860 (99.7\%) \\
- in S2ORC graph    & 260,014 / 311,860 (83.3\%) \\ 
\bottomrule
\end{tabular}
\end{table}

Neither the SPECTER training data nor the SciDocs test data comes with a mapping to the S2ORC dataset, which we use for the training of the citation embedding model.
However, to replicate SPECTER's training data and to avoid leakage of SciDocs test data such a mapping is needed.
Therefore, we try to map the papers to S2ORC based on PDF hashes and exact title matches. The remaining paper metadata is collected through the Semantic Scholar API.
\cref{tab:id-mapping} summarizes the outcome of mapping procedure.
Failed mappings can be attributed to papers being unavailable through the Semantic Scholar API (e.g., retracted  papers) or papers not being part of S2ORC citation graph.

\section{SPECTER-SciDocs Leakage} \label{ssec:leakage}

When replicating SPECTER \cite{Cohan2020}, we found a substantial overlap between the papers\footnote{\url{https://github.com/allenai/specter/issues/2}} used during the model training and the papers from their \dataset benchmark\footnote{\url{https://github.com/allenai/scidocs}}.
In both datasets, papers are associated with Semantic Scholar IDs. Thus, no custom ID mapping as in \cref{s2orc-mapping} is required to identify papers that leak from training to test data. 
From the 311,860 unique papers used in SPECTER's training data, we find 79,201 papers (25.4\%) in the test set of \dataset and 79,609 papers (25.5\%) in its validation set.
When combining test and validation set, there is a total overlap of 126,176 papers (40.5\%).
However, this overlap affects only the `unsupervised' paper metadata (title, abstract, citations, etc.) and not the gold labels used in \dataset (e.g., MAG labels or clicked recommendations).

\section{Dataset Creation} \label{ssec:dataset-creation}

As describe in \cref{ssec:training-data}, we conduct our experiments on two datasets. 
Both datasets rely on the citation graph of S2ORC \cite{Lo2020}.
More specifically, S2ORC with the version identifier \texttt{20200705v1} is used.
The full citation graph consists of 52.6M nodes (papers) and 467M edges (citations).
\cref{tab:dataset-stats} presents statistics on the datasets and their overlap with SPECTER and \dataset.
The steps to reproduce both datasets are: 

\paragraph{Replicated SPECTER (w/ leakage)}

In order to replicate SPECTER's training data and do not increase the leakage, we exclude all \dataset papers which are not used by SPECTER from the S2ORC citation graph.
This means that apart from the 110,538 SPECTER papers not a single other \dataset paper is included.
The resulting citation graph has 52.5M nodes and 463M edges and is used for training the citation graph embeddings.

For the \sys triplet selection, we also replicate SPECTER's query papers and its corpus from which positive and negatives are sampled. 
Our mapping and the underlying citation graph allows us to use 227,869 of 248,007 SPECTER's papers for training.
Regarding query papers, we use 131,644 of 136,820 SPECTER's query papers.
To align the number training triplets with the one from SPECTER, additional papers are randomly sampled from the filtered citation graph.

\paragraph{Random S2ORC subset (w/o leakage)}

To avoid leakage, we exclude all successfully mapped \dataset papers from the S2ORC citation graph.
After filtering the graph has 52.3 nodes and 447M edges.
The citation graph embedding model is trained on this graph.

Next, we reproduce triplet selection from SPECTER. 
Any random 136,820 query papers are selected from the filtered graph.
For each query, we generate five positives (cited by the query), two hard negatives (citation of citation), and three random nodes from the filtered S2ORC citation graphs.
This sampling produces 684,100 training triplets with 680,967 unique papers IDs (more compared to the replicated SPECTER dataset).
Based on these triplets the SPECTER model for this dataset is trained with the same model settings and hyperparameters as \sys (second last row in \cref{tab:results}).

Lastly, the \sys triplets are generated based on the citation graph embeddings of the same 680,967 unique papers IDs, i.e, the FAISS index contains only these papers and not the remaining S2ORC papers.
Also, the same 136,820 query papers are used.

\begin{table}[h]
\footnotesize
\centering
\caption{Statistics for our two datasets and their overlap with SPECTER and SciDocs respectively.}
\label{tab:dataset-stats}
\begin{tabular}{lrr}
\toprule
\multicolumn{1}{c}{\textbf{}} &
  \multicolumn{1}{c}{\textbf{\begin{tabular}[c]{@{}c@{}}Replicated \\SPECTER \\ (w/ leakage)\end{tabular}}} &
  \multicolumn{1}{c}{\textbf{\begin{tabular}[c]{@{}c@{}}Random \\S2ORC subset\\ (w/o leakage)\end{tabular}}} \\
  \midrule
Training triplets                                                        & 684,100    & 684,100    \\
\rule{0pt}{4ex}  

Unique paper IDs                                                        & 248,007    & 680,967    \\
- in SPECTER                                                            & 227,869    & 9,182      \\
- in SciDocs                                                            & 110,538    & 0         \\
\begin{tabular}[c]{@{}l@{}}- in SciDocs \\ ~~ and in SPECTER\end{tabular} & 110,538    & 0         \\

\rule{0pt}{4ex}  

Query paper IDs                                                         & 136,820    & 136,820    \\
- in SciDocs                                                            & 69,306     & 0         \\
- in SPECTER queries                                                   & 131,644    & 463       \\

\rule{0pt}{4ex}  

Citation graph                                                          &           &           \\
- Nodes                                                                 & 52,526,134  & 52,373,977  \\
- Edges                                                                 & 463,697,639 & 447,697,727 \\ 
\bottomrule
\end{tabular}
\end{table}

\section{Graph Embedding Evaluation} \label{ssec:graph-evaluation}

To evaluate the underlying citation graph embeddings, we experiment with a few of BigGraph's hyperparameters.
We trained embeddings with different dimensions $d{=}\{128,512,768\}$ and different distance measures (cosine similarity and dot product) on 99\% of the data and test the remaining 1\% on the link prediction task.
An evaluation of the graph embeddings with \dataset is not possible since we could not map the papers used in \dataset to the S2ORC corpus.
All variations are trained for 20 epochs, margin $m{=}0.15$, and learning rate $\lambda{=}0.1$ (based on the recommended settings by \citet{Lerer2019}).

\begin{table}[ht]
\centering
\footnotesize

\caption{Link prediction performance of BigGraph embeddings trained on S2ORC citation graph with different dimensions and distance measures.}
\label{tab:biggraph-results}
\begin{tabular}{ll|llll}
\toprule
\textbf{Dim.} & \textbf{Dist.} & \textbf{MRR} & \textbf{Hits@1} & \textbf{Hits@10} & \textbf{AUC} \\
\midrule
128 & Cos. & 54.09 & 43.39 & 75.21 & 85.75 \\
128 & Dot    & 89.75 & 85.84 & 96.13 & 97.70 \\
512 & Dot    & 94.60 & 92.47 & 97.64 & 98.64 \\
768 & Dot    & 95.12 & 93.22 & 97.77 & 98.74 \\ 
\bottomrule
\end{tabular}
\end{table}

\cref{tab:biggraph-results} shows the link prediction performance measured in MRR, Hits@1, Hits@10, and AUC.
Dot product is substantially better than cosine similarity as distance measure.
Also, there is a positive correlation between the performance and the size of the embeddings.
The larger the embedding size the better link prediction performance.
Graph embeddings with $d{=}768$ were the largest possible size given our compute resources (available disk space was the limiting factor).

\section{Baseline Details} \label{ssec:baseline-details}

If not otherwise mentioned, all BERT variations are used in their \textit{base-uncased} versions.

The weights for BERT (\textit{bert-base-uncased}), BioBERT (\textit{biobert-base-cased-v1.2}), CiteBERT (\textit{citebert}), DeCLUTR (\textit{declutr-sci-base}) are taken from Huggingface Hub\footnote{\url{https://huggingface.co/models}}.
We use Universal Sentence Encoder (USE) from Tensorflow Hub\footnote{\url{https://tfhub.dev/google/universal-sentence-encoder-large/5}}.
For \oracle, we use the SciNCL implementation and under-sample the triplets from the classification tasks to ensure a balanced triplet distribution over the tasks.
The SPECTER version for the random S2ORC training data (w/o leakage) is also trained with the SciNCL implementation.
Please see \citet{Cohan2020} for additional baseline methods and their implementation details.

\section{Negative Results} \label{ssec:negative-results}
We investigated additional sampling strategies and model modification of which none led to a significant performance improvement.

\subsection{Undirected Citations} \label{sssec:undirected-citations}
Our graph embedding model considers citations as directed edges by default.
We also train a \sys model with undirected citations by first converting a single edge $(a,b)$ into the two edges $(a,b)$ and $(b,a)$.
This approach yields a slightly worse performance (81.7 avg. score; -0.1 points) and, therefore, was discarded for the final experiments.

\subsection{\knn with interval large than $c$}
Our best results are achieved with \knn where the size of the neighbor interval $(k-c';k]$ is equal to the number of samples $c'$ that the strategy should generate.
In addition to this, we also experimented with large intervals, e.g., $(1000;2000]$, from which $c'$ papers are randomly sampled.
This approach yields comparable results but suffers from a larger effect of randomness and is therefore more difficult to optimize.

\subsection{K-Means Cluster for Easy Negatives} 
Easy negatives are supposed to be far away from the query.
Random sampling from a large corpus ensures this as our results show.
As an alternative approach, we tried k-means clustering whereby we selected easy negatives from the centroid that has a given distance to the query's centroid.
However, this decreased the performance.

\subsection{Sampling with Similarity Threshold}

As alternative to \knn, we select samples based on cosine similarity in the citation embedding space.
 Take $c'$ papers that are within the similarity threshold $t$ of a query paper $\query$ such that
 $s(f_c(\query),f_c(d_i))<t$, where $s$ is the cosine similarity function.

 For example, given the similarity scores $S{=}\{0.9,0.8,0.7,0.1\}$ (ascending order, the higher the similarity is the closer the candidate embedding to the query embedding is) with $c'{=}2$ and $t{=}0.5$, the two candidates with the largest similarity scores and larger than the threshold would be $0.8$ and $0.7$. 
 The corresponding papers would be selected as samples.
 While the positive threshold $t^+$ should close to 1, the negative threshold $t^-$ should be small  
 to ensure samples are dissimilar from $\query$.
However, the empirical results suggest that this strategy is inferior compared to \knn.

\subsection{Hard Negatives with Similarity Threshold}

\begin{figure}[ht]
\centering
\includegraphics[clip,width=1.\linewidth]{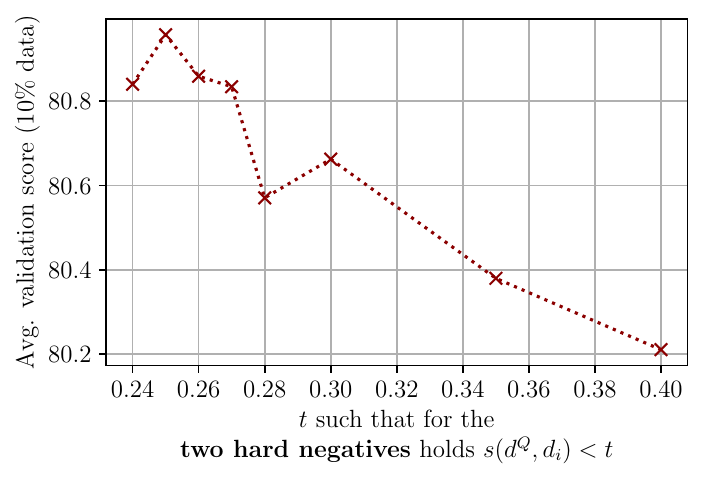}

\caption{\label{fig:hard_neg_range}Results on the validation set w.r.t. hard negative sampling with \simthreshold using 10\% training data.}
\end{figure}

Selecting hard negatives based on the similarity threshold yields a test score of  81.7 (-0.1 points).
\cref{fig:hard_neg_range} show the validation results for different similarity thresholds. 
A similar pattern as in \cref{fig:hard_neg} can be seen. 
When the negatives are closer to the query paper (larger similarity threshold $t$), the validation score decreases.

\subsection{Positives with Similarity Threshold}

Positive sampling with \simthreshold performs poorly since even for small $t^{+}<0.5$ many query papers do not have any neighbors within this similarity threshold (more than 40\%). 
Solving this issue would require changing the set of query papers which we omit for comparability to SPECTER.

\subsection{Sorted Random} 
Simple random sampling does not ensure if a sample is far or close to the query.
To integrate a distance measure in the random sampling, we first sample $n$ candidates, then order the candidates according to their distance to the query, and lastly select the $c'$ candidates that are the closest or furthest to the query as samples. 

\subsection{Mask Language Modeling}

\citet{Giorgi2021} show that combining a contrastive loss with a mask language modeling loss can improve text representation learning.
However, in our experiments a combined function decreases the performance on \dataset, probably due to the effects found by \cite{Li2020}. 

\subsection{Student-Teacher Learning}
\label{sssec:student-teacher}

Student-teacher learning is effective in related work on cross-modal knowledge transfer \citep{Kaur2021,Tian2020}. %
We also try to adopt this approach for our experiments, whereby the Transformer language model is the student, and the citation graph embedding model is the teacher.
By directly learning from the citation embeddings, we could circumvent the positive and negative sampling needed for triplet loss learning, which  introduces unwanted issues like collisions.
Given a batch of document representations derived from text ${D}_{Text}$ (through the language model) and the citation graph representations for the same documents $D_{Graph}$, we compute the pairwise cosine similarity for both sets ${S}_{Text}$ and ${S}_{Graph}$.
To transfer the knowledge from the citation embeddings into the language model, we devise the student-teacher loss $\calL_{ST}$ based on a mean-squared-error loss (MSE) such that the difference between the cosine similarities is minimized: 

\begin{equation}
    \calL_{ST}=\text{MSE}(S_{Text},S_{Graph})
\label{eq:st-loss}
\end{equation}

Despite the promising results from \citet{Tian2020}, the student-teacher approach performs poorly in our experiments.
We attribute this the overfitting to the citation data (the training loss approaches zero after a few steps while the validation loss remains high).
The model trained with $\calL_{ST}$ yields only a \dataset average score of 64.7, slightly better than SciBERT but substantially worse than \sys with triplet loss.

Additionally, we experiment with a joint loss that is the sum of triplet margin loss $\calL_{Triplet}$ (see \cref{ssec:cl-objective}) and the student-teacher loss $\calL_{ST}$:

\begin{equation}
    \calL_{Joint}=\calL_{Triplet}+\calL_{ST}
\label{eq:joint-loss}
\end{equation}

Training with the joint loss $\calL_{Joint}$ achieves an average score of 80.5. 
Even though the joint loss is not subject to overfitting, its \dataset performance is slightly worse than the triplet loss $\calL_{Triplet}$ alone.
Given this outcome and that the computation of the cosine similarities adds additional complexity, we discard the student-teacher approach for the final experiments.

\subsection{SPECTER \& Bidirectional Citations} \label{sssec:bi-specter}

SPECTER \citep{Cohan2020} relies on unidirectional citations for their sampling strategy.
While papers \textit{cited by} the query paper are considered as positives samples, those \textit{citing} the query paper (opposite citation direction) could be negative samples.
We see this use of citations as a conceptional flaw in their sampling strategy.

To test the actual effect on the resulting document representation, we first replicate the original unidirectional sampling strategy from SPECTER with our training data (see w/ leakage in \cref{ssec:training-data}). 
The resulting SPECTER model achieves an average score of 79.0 on \dataset.\footnote{The difference to the scores reported in \citet{Cohan2020} is due to the difference in the underlying training data.}
When changing the sampling strategy from unidirectional to bidirectional (`citations to the query' are also treated as a signal for similarity), we observe an improvement of +0.4 points to 79.4.
Consequently, the use of unidirectional citations is not only a conceptional issue but also degrades learning performance.

\section{Task-specific Results}

\begin{figure*}
     \centering
     \begin{subfigure}[b]{0.45\textwidth}
         \centering
         \includegraphics[width=\textwidth]{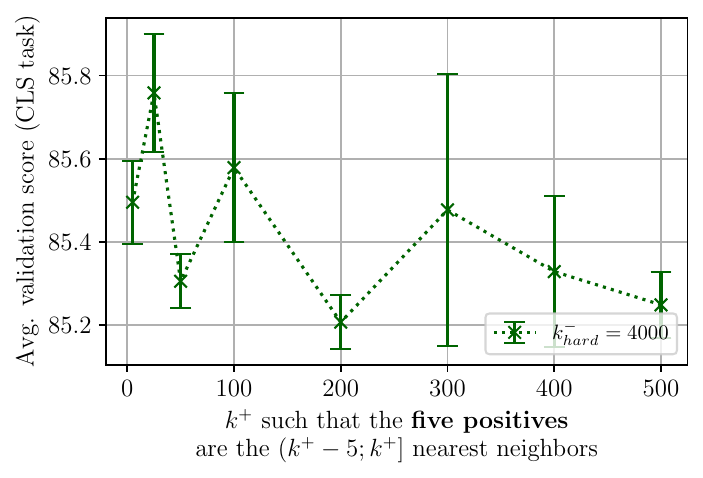}
         \caption{Classification}
         \label{fig:pos-cls}
     \end{subfigure}
     \hfill
     \begin{subfigure}[b]{0.45\textwidth}
         \centering
         \includegraphics[width=\textwidth]{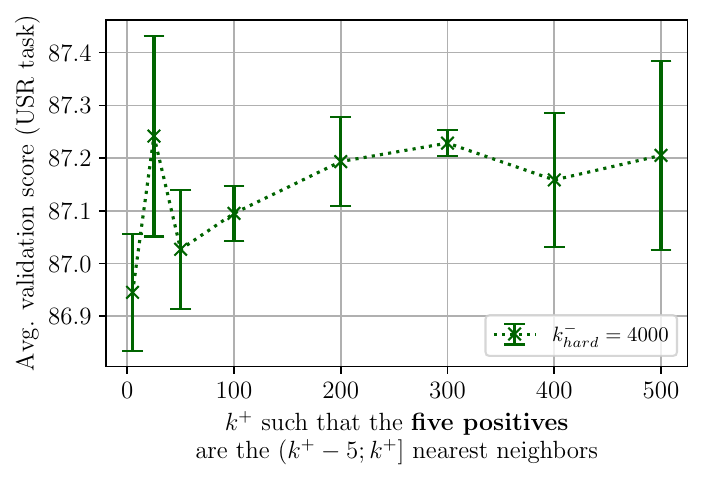}
         \caption{User activities}
         \label{fig:pos-usr}
     \end{subfigure}
     \hfill
     \begin{subfigure}[b]{0.45\textwidth}
         \centering
         \includegraphics[width=\textwidth]{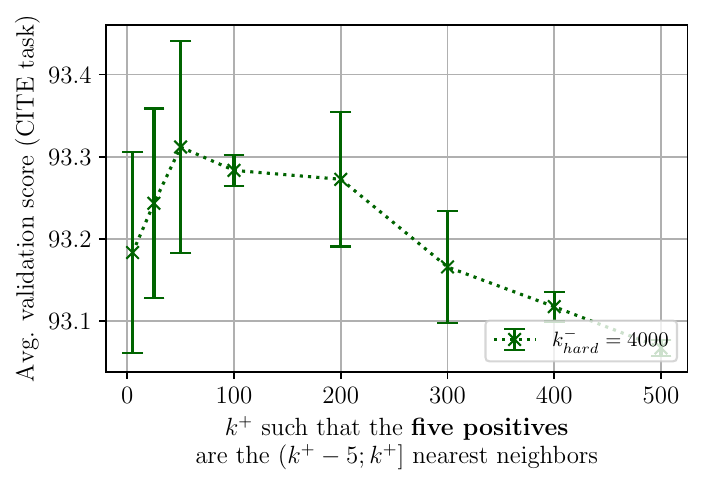}
         \caption{Citation}
         \label{fig:pos-cite}
     \end{subfigure}
     \hfill
     \begin{subfigure}[b]{0.45\textwidth}
         \centering
         \includegraphics[width=\textwidth]{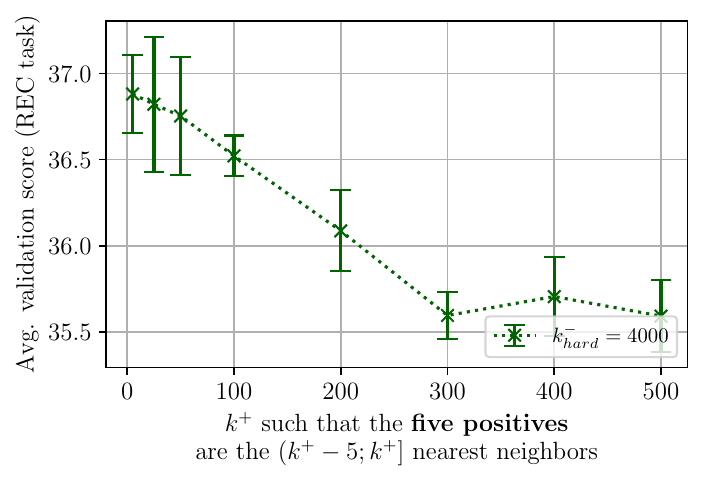}
         \caption{Recommendation}
         \label{fig:pos-rec}
     \end{subfigure}
        \caption{Task-level validation performance w.r.t. $\positiveK$ with \knn strategy using 10\% training data.}
        \label{fig:pos-tasks}
\end{figure*}

\begin{figure*}
     \centering
     \begin{subfigure}[b]{0.45\textwidth}
         \centering
         \includegraphics[width=\textwidth]{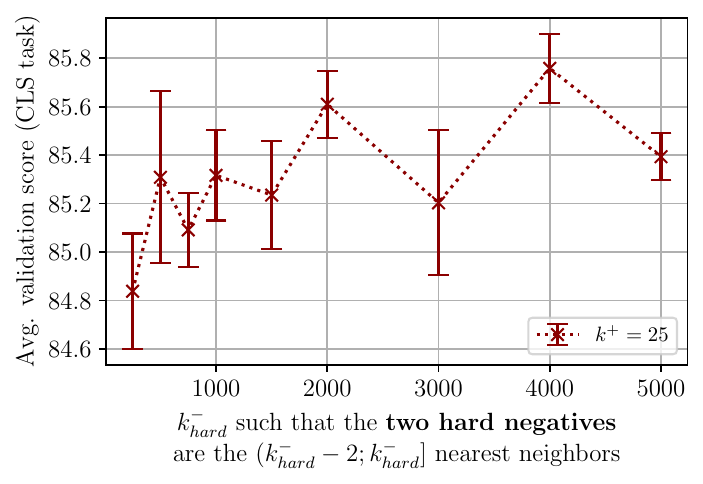}
         \caption{Classification}
         \label{fig:neg-cls}
     \end{subfigure}
     \hfill
     \begin{subfigure}[b]{0.45\textwidth}
         \centering
         \includegraphics[width=\textwidth]{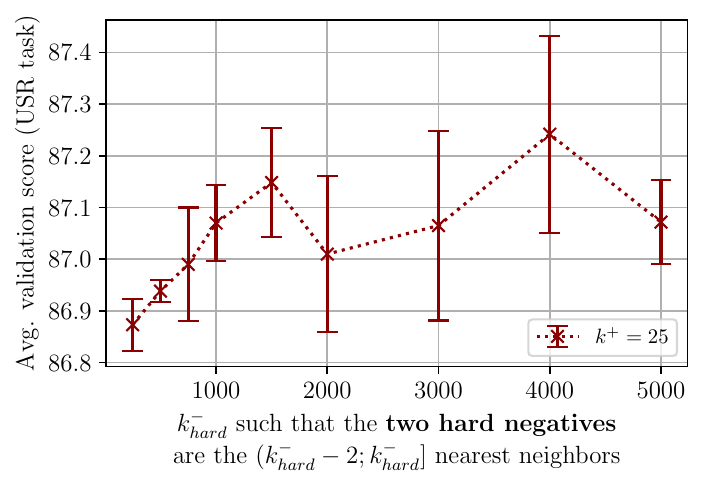}
         \caption{User activities}
         \label{fig:neg-usr}
     \end{subfigure}
     \hfill
     \begin{subfigure}[b]{0.45\textwidth}
         \centering
         \includegraphics[width=\textwidth]{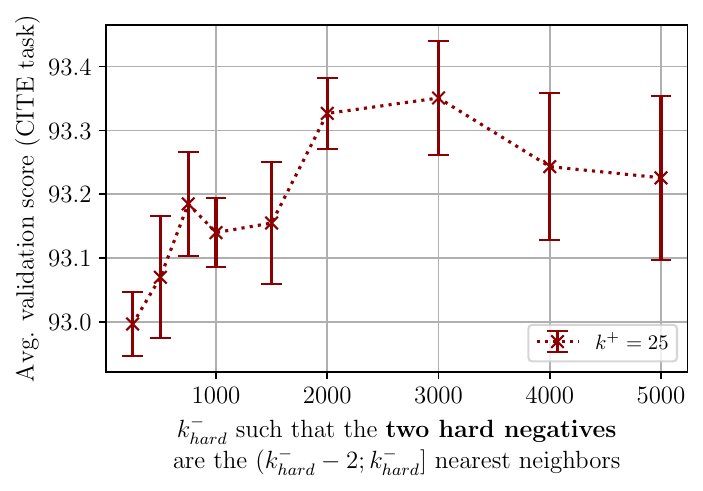}
         \caption{Citation}
         \label{fig:neg-cite}
     \end{subfigure}
     \hfill
     \begin{subfigure}[b]{0.45\textwidth}
         \centering
         \includegraphics[width=\textwidth]{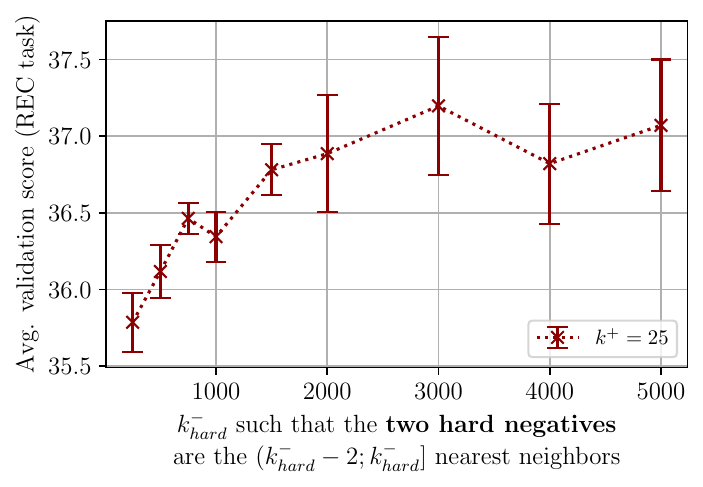}
         \caption{Recommendation}
         \label{fig:neg-rec}
     \end{subfigure}
        \caption{Task-level validation performance w.r.t. $\hardNegativeK$ with \knn strategy using 10\% training data.}
        \label{fig:neg-tasks}
\end{figure*}

\cref{fig:pos-tasks} and \ref{fig:neg-tasks} present the validation performance like in \cref{sec:hyperparameter-analysis} but on a task-level and not as an average over all tasks. 
The plots show that the optimal $\positiveK$ and $\hardNegativeK$ values are partially task dependent.

\section{Examples} \label{ssec:examples}

\cref{tab:examples} lists three examples of query papers with their corresponding positive and negative samples.
The complete set of triplets that we use during training is available in our code repository\footref{fn:github}.

\begin{table*}[ht]
\centering

\footnotesize
\caption{Example query papers with their positive and negative samples.}
\label{tab:examples}
\begin{tabular}{ll}
\toprule \\

Query: & \textbf{BERT: Pre-training of Deep Bidirectional Transformers for Language Understanding} \\ \rule{0pt}{8ex} Positives: & \begin{tabular}[c]{@{}l@{}} 
~~\llap{\textbullet}~~ A Broad-Coverage Challenge Corpus for Sentence Understanding through Inference\\
~~\llap{\textbullet}~~ Looking for ELMo's Friends: Sentence-Level Pretraining Beyond Language Modeling\\
~~\llap{\textbullet}~~ GLUE : A MultiTask Benchmark and Analysis Platform for Natural Language Understanding\\
~~\llap{\textbullet}~~ Dissecting Contextual Word Embeddings: Architecture and Representation\\
~~\llap{\textbullet}~~ Universal Transformers\\
\end{tabular} \\ \rule{0pt}{8ex} Negatives: & \begin{tabular}[c]{@{}l@{}}~~\llap{\textbullet}~~ Planning for decentralized control of multiple robots under uncertainty\\
~~\llap{\textbullet}~~ Graph-Based Relational Data Visualization\\
~~\llap{\textbullet}~~ Linked Stream Data Processing\\
~~\llap{\textbullet}~~ Topic Modeling Using Distributed Word Embeddings\\
~~\llap{\textbullet}~~ Adversarially-Trained Normalized Noisy-Feature Auto-Encoder for Text Generation\\
\end{tabular}   \\
\rule{0pt}{3ex}   \\
\hline   
\rule{0pt}{5ex}

Query: & \textbf{BioBERT: a pre-trained biomedical language representation model for biomedical text mining} \\ \rule{0pt}{8ex} Positives: & \begin{tabular}[c]{@{}l@{}} 
~~\llap{\textbullet}~~ Exploring Word Embedding for Drug Name Recognition\\
~~\llap{\textbullet}~~ A neural joint model for entity and relation extraction from biomedical text\\
~~\llap{\textbullet}~~ Event Detection with Hybrid Neural Architecture\\
~~\llap{\textbullet}~~ Improving chemical disease relation extraction with rich features and weakly labeled data\\
~~\llap{\textbullet}~~ GLUE : A MultiTask Benchmark and Analysis Platform for Natural Language Understanding\\
\end{tabular} \\ 

\rule{0pt}{8ex} 

Negatives: & \begin{tabular}[c]{@{}l@{}}~~\llap{\textbullet}~~ Weakly Supervised Facial Attribute Manipulation via Deep Adversarial Network\\
~~\llap{\textbullet}~~ Applying the Clique Percolation Method to analyzing cross-market branch banking ...\\
~~\llap{\textbullet}~~ Perpetual environmentally powered sensor networks\\
~~\llap{\textbullet}~~ Labelling strategies for hierarchical multi-label classification techniques\\
~~\llap{\textbullet}~~ Domain Aware Neural Dialog System\\
\end{tabular}  \\ \rule{0pt}{3ex}   \\
\hline   
\rule{0pt}{5ex}

Query: & \textbf{A Context-Aware Citation Recommendation Model with BERT and Graph Convolutional Networks} \\ \rule{0pt}{8ex} Positives: & \begin{tabular}[c]{@{}l@{}} 
~~\llap{\textbullet}~~ Content-based citation analysis: The next generation of citation analysis\\
~~\llap{\textbullet}~~ ScisummNet: A Large Annotated Dataset and Content-Impact Models for Scientific Paper ...\\
~~\llap{\textbullet}~~ Citation Block Determination Using Textual Coherence\\
~~\llap{\textbullet}~~ Discourse Segmentation Of Multi-Party Conversation\\
~~\llap{\textbullet}~~ Argumentative Zoning for Improved Citation Indexing\\
\end{tabular} \\ \rule{0pt}{8ex} Negatives: & \begin{tabular}[c]{@{}l@{}}~~\llap{\textbullet}~~ Adaptive Quantization for Hashing: An Information-Based Approach to Learning ...\\
~~\llap{\textbullet}~~ Trap Design for Vibratory Bowl Feeders\\
~~\llap{\textbullet}~~ Software system for the Mars 2020 mission sampling and caching testbeds\\
~~\llap{\textbullet}~~ Applications of Rhetorical Structure Theory\\
~~\llap{\textbullet}~~ Text summarization for Malayalam documents — An experience\\
\end{tabular}   \\  
                                           
\bottomrule
\end{tabular}
\end{table*}

\end{document}

%% file: IJCAI_math_commands.tex
\usepackage{amsmath,amsfonts,bm}

\def\eqref#1{equation~\ref{#1}}

\def\1{\bm{1}}

\def\vd{{\bm{d}}}

\DeclareMathAlphabet{\mathsfit}{\encodingdefault}{\sfdefault}{m}{sl}
\SetMathAlphabet{\mathsfit}{bold}{\encodingdefault}{\sfdefault}{bx}{n}

%% file: paper.bbl
\begin{thebibliography}{61}
\expandafter\ifx\csname natexlab\endcsname\relax\def\natexlab#1{#1}\fi

\bibitem[{Beltagy et~al.(2019)Beltagy, Lo, and Cohan}]{Beltagy2019}
Iz~Beltagy, Kyle Lo, and Arman Cohan. 2019.
\newblock \href {https://doi.org/10.18653/v1/D19-1371} {{SciBERT: A Pretrained
  Language Model for Scientific Text}}.
\newblock In \emph{Proceedings of the 2019 Conference on Empirical Methods in
  Natural Language Processing and the 9th International Joint Conference on
  Natural Language Processing (EMNLP-IJCNLP)}, pages 3613--3618, Stroudsburg,
  PA, USA. Association for Computational Linguistics.

\bibitem[{Ben~Zaken et~al.(2022)Ben~Zaken, Goldberg, and Ravfogel}]{Zaken2021}
Elad Ben~Zaken, Yoav Goldberg, and Shauli Ravfogel. 2022.
\newblock \href {https://doi.org/10.18653/v1/2022.acl-short.1} {{B}it{F}it:
  Simple parameter-efficient fine-tuning for transformer-based masked
  language-models}.
\newblock In \emph{Proceedings of the 60th Annual Meeting of the Association
  for Computational Linguistics (Volume 2: Short Papers)}, pages 1--9, Dublin,
  Ireland. Association for Computational Linguistics.

\bibitem[{Bhagavatula et~al.(2018)Bhagavatula, Feldman, Power, and
  Ammar}]{Bhagavatula2018}
Chandra Bhagavatula, Sergey Feldman, Russell Power, and Waleed Ammar. 2018.
\newblock \href {https://doi.org/10.18653/v1/n18-1022} {{Content-based citation
  recommendation}}.
\newblock \emph{NAACL HLT 2018 - 2018 Conference of the North American Chapter
  of the Association for Computational Linguistics: Human Language Technologies
  - Proceedings of the Conference}, 1:238--251.

\bibitem[{Brochier et~al.(2019)Brochier, Guille, and Velcin}]{Brochier2019}
Robin Brochier, Adrien Guille, and Julien Velcin. 2019.
\newblock \href {https://doi.org/10.1145/3308558.3313595} {{Global Vectors for
  Node Representations}}.
\newblock In \emph{The World Wide Web Conference on - WWW '19}, volume~2, pages
  2587--2593, New York, New York, USA. ACM Press.

\bibitem[{Bucher et~al.(2016)Bucher, Herbin, and Jurie}]{Bucher2016}
Maxime Bucher, St{\'e}phane Herbin, and Fr{\'e}d{\'e}ric Jurie. 2016.
\newblock \href {https://arxiv.org/abs/1608.07441} {Hard negative mining for
  metric learning based zero-shot classification}.
\newblock In \emph{Computer Vision -- ECCV 2016 Workshops}, pages 524--531,
  Cham. Springer International Publishing.

\bibitem[{Cer et~al.(2018)Cer, Yang, yi~Kong, Hua, Limtiaco, John, Constant,
  Guajardo-Cespedes, Yuan, Tar, Sung, Strope, and
  Kurzweil}]{Cer2018UniversalSE}
Daniel~Matthew Cer, Yinfei Yang, Sheng yi~Kong, Nan Hua, Nicole Limtiaco,
  Rhomni~St. John, Noah Constant, Mario Guajardo-Cespedes, Steve Yuan, Chris
  Tar, Yun-Hsuan Sung, Brian Strope, and Ray Kurzweil. 2018.
\newblock \href {http://arxiv.org/abs/1803.11175} {Universal sentence encoder}.
\newblock \emph{arXiv:1803.11175}.

\bibitem[{Cohan et~al.(2019)Cohan, Ammar, van Zuylen, and Cady}]{Cohan2019}
Arman Cohan, Waleed Ammar, Madeleine van Zuylen, and Field Cady. 2019.
\newblock \href {https://doi.org/10.18653/v1/N19-1361} {{Structural Scaffolds
  for Citation Intent Classification in Scientific Publications}}.
\newblock In \emph{Proceedings of the 2019 Conference of the North}, volume~1,
  pages 3586--3596, Stroudsburg, PA, USA. Association for Computational
  Linguistics.

\bibitem[{Cohan et~al.(2020)Cohan, Feldman, Beltagy, Downey, and
  Weld}]{Cohan2020}
Arman Cohan, Sergey Feldman, Iz~Beltagy, Doug Downey, and Daniel Weld. 2020.
\newblock \href {https://doi.org/10.18653/v1/2020.acl-main.207} {{SPECTER:
  Document-level Representation Learning using Citation-informed
  Transformers}}.
\newblock In \emph{Proceedings of the 58th Annual Meeting of the Association
  for Computational Linguistics}, pages 2270--2282, Stroudsburg, PA, USA.
  Association for Computational Linguistics.

\bibitem[{Devlin et~al.(2019)Devlin, Chang, Lee, and Toutanova}]{Devlin2019}
Jacob Devlin, Ming-Wei Chang, Kenton Lee, and Kristina Toutanova. 2019.
\newblock \href {https://doi.org/10.18653/v1/N19-1423} {{BERT: Pre-training of
  Deep Bidirectional Transformers for Language Understanding}}.
\newblock In \emph{Proceedings of the 2019 Conference of the North American
  Chapter of the Association for Computational Linguistics}, pages 4171--4186,
  Minneapolis, Minnesota. Association for Computational Linguistics.

\bibitem[{Elkiss et~al.(2008)Elkiss, Shen, Fader, Erkan, States, and
  Radev}]{Elkiss2008}
Aaron Elkiss, Siwei Shen, Anthony Fader, G{\"{u}}neş Erkan, David States, and
  Dragomir Radev. 2008.
\newblock \href {https://doi.org/10.1002/asi.20707} {{Blind men and elephants:
  What do citation summaries tell us about a research article?}}
\newblock \emph{Journal of the American Society for Information Science and
  Technology}, 59(1):51--62.

\bibitem[{Fang et~al.(2020)Fang, Wang, Zhou, Ding, and Xie}]{Fang2020}
Hongchao Fang, Sicheng Wang, Meng Zhou, Jiayuan Ding, and Pengtao Xie. 2020.
\newblock \href {https://doi.org/10.36227/techrxiv.12308378} {{CERT:
  Contrastive Self-supervised Learning for Language Understanding}}.
\newblock \emph{arXiv:2005.12766}, pages 1--16.

\bibitem[{Gao et~al.(2021)Gao, Yao, and Chen}]{Gao2021}
Tianyu Gao, Xingcheng Yao, and Danqi Chen. 2021.
\newblock \href {https://doi.org/10.18653/v1/2021.emnlp-main.552} {{S}im{CSE}:
  Simple contrastive learning of sentence embeddings}.
\newblock In \emph{Proceedings of the 2021 Conference on Empirical Methods in
  Natural Language Processing}, pages 6894--6910, Online and Punta Cana,
  Dominican Republic. Association for Computational Linguistics.

\bibitem[{Giorgi et~al.(2021)Giorgi, Nitski, Wang, and Bader}]{Giorgi2021}
John Giorgi, Osvald Nitski, Bo~Wang, and Gary Bader. 2021.
\newblock \href {https://doi.org/10.18653/v1/2021.acl-long.72} {{DeCLUTR: Deep
  Contrastive Learning for Unsupervised Textual Representations}}.
\newblock In \emph{Proceedings of the 59th Annual Meeting of the Association
  for Computational Linguistics and the 11th International Joint Conference on
  Natural Language Processing (Volume 1: Long Papers)}, pages 879--895,
  Stroudsburg, PA, USA. Association for Computational Linguistics.

\bibitem[{Gipp and Beel(2009)}]{Gipp2009}
Bela Gipp and J~Beel. 2009.
\newblock {Citation Proximity Analysis (CPA) - A new approach for identifying
  related work based on Co-Citation Analysis}.
\newblock \emph{Birger Larsen and Jacqueline Leta, editors, Proceedings of the
  12th International Conference on Scientometrics and Informetrics (ISSI'09)},
  2(July):571--575.

\bibitem[{Grover and Leskovec(2016)}]{Grover2016}
Aditya Grover and Jure Leskovec. 2016.
\newblock \href {https://doi.org/10.1145/2939672.2939754} {{node2vec: Scalable
  Feature Learning for Networks}}.
\newblock In \emph{Proceedings of the 22nd ACM SIGKDD International Conference
  on Knowledge Discovery and Data Mining - KDD '16}, pages 855--864, New York,
  New York, USA. ACM Press.

\bibitem[{Han et~al.(2018)Han, Song, Zhao, Shi, and Zhang}]{Han2018}
Jialong Han, Yan Song, Wayne~Xin Zhao, Shuming Shi, and Haisong Zhang. 2018.
\newblock \href {https://doi.org/10.18653/v1/P18-1222} {{hyperdoc2vec:
  Distributed Representations of Hypertext Documents}}.
\newblock In \emph{Proceedings of the 56th Annual Meeting of the Association
  for Computational Linguistics (Volume 1: Long Papers)}, volume~1, pages
  2384--2394, Stroudsburg, PA, USA. Association for Computational Linguistics.

\bibitem[{Holm et~al.(2022)Holm, Plank, Wright, and
  Augenstein}]{holm2022longitudinal}
Andreas~Nugaard Holm, Barbara Plank, Dustin Wright, and Isabelle Augenstein.
  2022.
\newblock {Longitudinal Citation Prediction using Temporal Graph Neural
  Networks}.
\newblock In \emph{AAAI 2022 Workshop on Scientific Document Understanding (SDU
  2022)}.

\bibitem[{Jeong et~al.(2020)Jeong, Jang, Shin, Park, and Choi}]{Jeong2020ACC}
Chanwoo Jeong, Sion Jang, Hyuna Shin, Eunjeong~Lucy Park, and Sungchul Choi.
  2020.
\newblock \href {https://doi.org/10.1007/s11192-020-03561-y} {A context-aware
  citation recommendation model with bert and graph convolutional networks}.
\newblock \emph{Scientometrics}, pages 1--16.

\bibitem[{Johnson et~al.(2021)Johnson, Douze, and Jegou}]{Johnson2021}
Jeff Johnson, Matthijs Douze, and Herve Jegou. 2021.
\newblock \href {https://doi.org/10.1109/TBDATA.2019.2921572} {{Billion-Scale
  Similarity Search with GPUs}}.
\newblock \emph{IEEE Transactions on Big Data}, 7(3):535--547.

\bibitem[{Kaur et~al.(2021)Kaur, Pannu, and Malhi}]{Kaur2021}
Parminder Kaur, Husanbir~Singh Pannu, and Avleen~Kaur Malhi. 2021.
\newblock \href {https://doi.org/https://doi.org/10.1016/j.cosrev.2020.100336}
  {Comparative analysis on cross-modal information retrieval: A review}.
\newblock \emph{Computer Science Review}, 39:100336.

\bibitem[{Kessler(1963)}]{Kessler1963}
M.~M. Kessler. 1963.
\newblock \href {https://doi.org/10.1002/asi.5090140103} {{Bibliographic
  coupling between scientific papers}}.
\newblock \emph{American Documentation}, 14(1):10--25.

\bibitem[{Khosla et~al.(2020)Khosla, Teterwak, Wang, Sarna, Tian, Isola,
  Maschinot, Liu, and Krishnan}]{muliplePositiveCTL1}
Prannay Khosla, Piotr Teterwak, Chen Wang, Aaron Sarna, Yonglong Tian, Phillip
  Isola, Aaron Maschinot, Ce~Liu, and Dilip Krishnan. 2020.
\newblock \href
  {https://proceedings.neurips.cc/paper/2020/file/d89a66c7c80a29b1bdbab0f2a1a94af8-Paper.pdf}
  {Supervised contrastive learning}.
\newblock In \emph{Advances in Neural Information Processing Systems},
  volume~33, pages 18661--18673. Curran Associates, Inc.

\bibitem[{Kim et~al.(2021)Kim, Yoo, and Lee}]{Kim2021}
Taeuk Kim, Kang~Min Yoo, and Sang-goo Lee. 2021.
\newblock \href {https://doi.org/10.18653/v1/2021.acl-long.197} {Self-guided
  contrastive learning for {BERT} sentence representations}.
\newblock In \emph{Proceedings of the 59th Annual Meeting of the Association
  for Computational Linguistics and the 11th International Joint Conference on
  Natural Language Processing (Volume 1: Long Papers)}, pages 2528--2540,
  Online. Association for Computational Linguistics.

\bibitem[{Kingma and Ba(2015)}]{Kingma2015}
Diederik~P. Kingma and Jimmy~Lei Ba. 2015.
\newblock \href {http://arxiv.org/abs/1412.6980} {{Adam: A method for
  stochastic optimization}}.
\newblock \emph{3rd International Conference on Learning Representations, ICLR
  2015 - Conference Track Proceedings}, pages 1--15.

\bibitem[{Lee et~al.(2019)Lee, Yoon, Kim, Kim, Kim, So, and Kang}]{Lee2019}
Jinhyuk Lee, Wonjin Yoon, Sungdong Kim, Donghyeon Kim, Sunkyu Kim, Chan~Ho So,
  and Jaewoo Kang. 2019.
\newblock \href {https://doi.org/10.1093/bioinformatics/btz682} {{BioBERT: a
  pre-trained biomedical language representation model for biomedical text
  mining}}.
\newblock \emph{Bioinformatics}, pages 1--8.

\bibitem[{Lerer et~al.(2019)Lerer, Wu, Shen, Lacroix, Wehrstedt, Bose, and
  Peysakhovich}]{Lerer2019}
Adam Lerer, Ledell Wu, Jiajun Shen, Timothee Lacroix, Luca Wehrstedt, Abhijit
  Bose, and Alex Peysakhovich. 2019.
\newblock \href {http://arxiv.org/abs/1903.12287} {{PyTorch-BigGraph: A
  Large-scale Graph Embedding System}}.
\newblock In \emph{Proceedings of The Conference on Systems and Machine
  Learning}.

\bibitem[{Li et~al.(2020)Li, Zhou, He, Wang, Yang, and Li}]{Li2020}
Bohan Li, Hao Zhou, Junxian He, Mingxuan Wang, Yiming Yang, and Lei Li. 2020.
\newblock \href {https://doi.org/10.18653/v1/2020.emnlp-main.733} {{On the
  Sentence Embeddings from Pre-trained Language Models}}.
\newblock In \emph{Proceedings of the 2020 Conference on Empirical Methods in
  Natural Language Processing (EMNLP)}, pages 9119--9130, Stroudsburg, PA, USA.
  Association for Computational Linguistics.

\bibitem[{Lipscomb(2000)}]{Lipscomb2000MedicalSH}
Carolyn~E. Lipscomb. 2000.
\newblock Medical subject headings (mesh).
\newblock \emph{Bulletin of the Medical Library Association}, 88 3:265--6.

\bibitem[{Lo et~al.(2020)Lo, Wang, Neumann, Kinney, and Weld}]{Lo2020}
Kyle Lo, Lucy~Lu Wang, Mark Neumann, Rodney Kinney, and Daniel Weld. 2020.
\newblock \href {https://doi.org/10.18653/v1/2020.acl-main.447} {{S2ORC: The
  Semantic Scholar Open Research Corpus}}.
\newblock In \emph{Proceedings of the 58th Annual Meeting of the Association
  for Computational Linguistics}, pages 4969--4983, Stroudsburg, PA, USA.
  Association for Computational Linguistics.

\bibitem[{Loshchilov and Hutter(2019)}]{Loshchilov2019}
Ilya Loshchilov and Frank Hutter. 2019.
\newblock \href {http://arxiv.org/abs/1711.05101} {{Decoupled weight decay
  regularization}}.
\newblock \emph{7th International Conference on Learning Representations, ICLR
  2019}.

\bibitem[{Luu et~al.(2021)Luu, Wu, Koncel-Kedziorski, Lo, Cachola, and
  Smith}]{Luu2021}
Kelvin Luu, Xinyi Wu, Rik Koncel-Kedziorski, Kyle Lo, Isabel Cachola, and
  Noah~A. Smith. 2021.
\newblock \href {https://doi.org/10.18653/v1/2021.acl-long.166} {{Explaining
  Relationships Between Scientific Documents}}.
\newblock In \emph{Proceedings of the 59th Annual Meeting of the Association
  for Computational Linguistics and the 11th International Joint Conference on
  Natural Language Processing (Volume 1: Long Papers)}, pages 2130--2144,
  Stroudsburg, PA, USA. Association for Computational Linguistics.

\bibitem[{Musgrave et~al.(2020)Musgrave, Belongie, and
  Lim}]{MetricLearningReality}
Kevin Musgrave, Serge~J. Belongie, and Ser{-}Nam Lim. 2020.
\newblock \href {https://doi.org/10.1007/978-3-030-58595-2\_41} {A metric
  learning reality check}.
\newblock In \emph{Computer Vision - {ECCV} 2020 - 16th European Conference,
  Glasgow, UK, August 23-28, 2020, Proceedings, Part {XXV}}, pages 681--699.

\bibitem[{Ostendorff et~al.(2020)Ostendorff, Ruas, Blume, Gipp, and
  Rehm}]{Ostendorff2020c}
Malte Ostendorff, Terry Ruas, Till Blume, Bela Gipp, and Georg Rehm. 2020.
\newblock \href {http://arxiv.org/abs/2010.06395} {{Aspect-based Document
  Similarity for Research Papers}}.
\newblock In \emph{Proceedings of the 28th International Conference on
  Computational Linguistics (COLING 2020)}.

\bibitem[{Pasternack(1969)}]{Pasternack1969}
Simon Pasternack. 1969.
\newblock \href {https://doi.org/10.1126/science.164.3880.669} {The scientific
  enterprise: Public knowledge. an essay concerning the social dimension of
  science}.
\newblock \emph{Science}, 164(3880):669--670.

\bibitem[{Perozzi et~al.(2014)Perozzi, Al-Rfou, and Skiena}]{Perozzi2014}
Bryan Perozzi, Rami Al-Rfou, and Steven Skiena. 2014.
\newblock \href {https://doi.org/10.1145/2623330.2623732} {{DeepWalk: online
  learning of social representations}}.
\newblock In \emph{Proceedings of the 20th ACM SIGKDD international conference
  on Knowledge discovery and data mining - KDD '14}, pages 701--710, New York,
  New York, USA. ACM Press.

\bibitem[{Rehm et~al.(2020)Rehm, Bourgonje, Hegele, Kintzel, Schneider,
  Ostendorff, Zaczynska, Berger, Grill, Räuchle, Rauenbusch, Rutenburg,
  Schmidt, Wild, Hoffmann, Fink, Schulz, Seva, Quantz, Böttger, Matthey,
  Fricke, Thomsen, Paschke, Qundus, Hoppe, Karam, Weichhardt, Fillies,
  Neudecker, Gerber, Labusch, Rezanezhad, Schaefer, Zellhöfer, Siewert, Bunk,
  Pintscher, Aleynikova, and Heine}]{rehm2020d}
Georg Rehm, Peter Bourgonje, Stefanie Hegele, Florian Kintzel, Julán~Moreno
  Schneider, Malte Ostendorff, Karolina Zaczynska, Armin Berger, Stefan Grill,
  Sören Räuchle, Jens Rauenbusch, Lisa Rutenburg, André Schmidt, Mikka Wild,
  Henry Hoffmann, Julian Fink, Sarah Schulz, Jurica Seva, Joachim Quantz,
  Joachim Böttger, Josefine Matthey, Rolf Fricke, Jan Thomsen, Adrian Paschke,
  Jamal~Al Qundus, Thomas Hoppe, Naouel Karam, Frauke Weichhardt, Christian
  Fillies, Clemens Neudecker, Mike Gerber, Kai Labusch, Vahid Rezanezhad, Robin
  Schaefer, David Zellhöfer, Daniel Siewert, Patrick Bunk, Lydia Pintscher,
  Elena Aleynikova, and Franziska Heine. 2020.
\newblock {QURATOR: Innovative Technologies for Content and Data Curation}.
\newblock In \emph{Proceedings of QURATOR 2020 -- The conference for
  intelligent content solutions}, Berlin, Germany.
\newblock CEUR Workshop Proceedings, Volume 2535. 20/21 January 2020.

\bibitem[{Reimers and Gurevych(2019)}]{Reimers2019}
Nils Reimers and Iryna Gurevych. 2019.
\newblock \href {https://doi.org/10.18653/v1/D19-1410} {{Sentence-BERT:
  Sentence Embeddings using Siamese BERT-Networks}}.
\newblock In \emph{Proceedings of the 2019 Conference on Empirical Methods in
  Natural Language Processing and the 9th International Joint Conference on
  Natural Language Processing (EMNLP-IJCNLP)}, pages 3980--3990, Stroudsburg,
  PA, USA. Association for Computational Linguistics.

\bibitem[{Rethmeier and Augenstein(2021)}]{rethmeier2021dataefficient}
Nils Rethmeier and Isabelle Augenstein. 2021.
\newblock \href {http://arxiv.org/abs/2010.01061} {{Data-Efficient Pretraining
  via Contrastive Self-Supervision}}.
\newblock \emph{arXiv:2102.12982}.

\bibitem[{Rethmeier and Augenstein(2022)}]{Rethmeier2021}
Nils Rethmeier and Isabelle Augenstein. 2022.
\newblock \href {https://doi.org/10.1145/3561970} {A primer on contrastive
  pretraining in language processing: Methods, lessons learned \&
  perspectives}.
\newblock \emph{ACM Comput. Surv.}

\bibitem[{Rogers et~al.(2020)Rogers, Kovaleva, and Rumshisky}]{BERTOLOGY}
Anna Rogers, Olga Kovaleva, and Anna Rumshisky. 2020.
\newblock \href {https://transacl.org/ojs/index.php/tacl/article/view/2257} {A
  primer in bertology: What we know about how {BERT} works}.
\newblock \emph{Trans. Assoc. Comput. Linguistics}, 8:842--866.

\bibitem[{Saunshi et~al.(2019)Saunshi, Plevrakis, Arora, Khodak, and
  Khandeparkar}]{ContrastiveLearningLimitations}
Nikunj Saunshi, Orestis Plevrakis, Sanjeev Arora, Mikhail Khodak, and
  Hrishikesh Khandeparkar. 2019.
\newblock \href {http://proceedings.mlr.press/v97/saunshi19a.html} {A
  theoretical analysis of contrastive unsupervised representation learning}.
\newblock In \emph{{ICML}}, volume~97 of \emph{{PMLR}}. PMLR.

\bibitem[{Schroff et~al.(2015)Schroff, Kalenichenko, and
  Philbin}]{TripletLossORG}
Florian Schroff, Dmitry Kalenichenko, and James Philbin. 2015.
\newblock \href {https://doi.org/10.1109/CVPR.2015.7298682} {Facenet: {A}
  unified embedding for face recognition and clustering}.
\newblock In \emph{{IEEE} Conference on Computer Vision and Pattern
  Recognition, {CVPR} 2015, Boston, MA, USA, June 7-12, 2015}, pages 815--823.

\bibitem[{Shorten et~al.(2021)Shorten, Khoshgoftaar, and
  Furht}]{TextAugemntations}
Connor Shorten, Taghi~M. Khoshgoftaar, and Borko Furht. 2021.
\newblock \href {https://doi.org/10.1186/s40537-021-00492-0} {Text data
  augmentation for deep learning}.
\newblock \emph{J. Big Data}, 8(1):101.

\bibitem[{Sinha et~al.(2015)Sinha, Shen, Song, Ma, Eide, Hsu, and
  Wang}]{Sinha2015AnOO}
Arnab Sinha, Zhihong Shen, Yang Song, Hao Ma, Darrin Eide, Bo-June~Paul Hsu,
  and Kuansan Wang. 2015.
\newblock \href {https://doi.org/10.1145/2740908.2742839} {An overview of
  microsoft academic service (mas) and applications}.
\newblock \emph{Proceedings of the 24th International Conference on World Wide
  Web}.

\bibitem[{Small(1973)}]{Small1973}
Henry Small. 1973.
\newblock \href {https://doi.org/10.1002/asi.4630240406} {{Co-citation in the
  scientific literature: A new measure of the relationship between two
  documents}}.
\newblock \emph{Journal of the American Society for Information Science},
  24(4):265--269.

\bibitem[{Socher et~al.(2013)Socher, Ganjoo, Manning, and Ng}]{Socher2013}
Richard Socher, Milind Ganjoo, Christopher~D Manning, and Andrew Ng. 2013.
\newblock \href
  {https://proceedings.neurips.cc/paper/2013/file/2d6cc4b2d139a53512fb8cbb3086ae2e-Paper.pdf}
  {Zero-shot learning through cross-modal transfer}.
\newblock In \emph{Advances in Neural Information Processing Systems},
  volume~26. Curran Associates, Inc.

\bibitem[{Teufel et~al.(2006)Teufel, Siddharthan, and Tidhar}]{Teufel2006}
Simone Teufel, Advaith Siddharthan, and Dan Tidhar. 2006.
\newblock \href {https://doi.org/10.3115/1610075.1610091} {{Automatic
  classification of citation function}}.
\newblock In \emph{Proceedings of the 2006 Conference on Empirical Methods in
  Natural Language Processing - EMNLP '06}, page 103, Morristown, NJ, USA.
  Association for Computational Linguistics.

\bibitem[{Tian et~al.(2020{\natexlab{a}})Tian, Krishnan, and Isola}]{Tian2020}
Yonglong Tian, Dilip Krishnan, and Phillip Isola. 2020{\natexlab{a}}.
\newblock \href {https://openreview.net/forum?id=SkgpBJrtvS} {Contrastive
  representation distillation}.
\newblock In \emph{International Conference on Learning Representations}.

\bibitem[{Tian et~al.(2020{\natexlab{b}})Tian, Sun, Poole, Krishnan, Schmid,
  and Isola}]{MI_views}
Yonglong Tian, Chen Sun, Ben Poole, Dilip Krishnan, Cordelia Schmid, and
  Phillip Isola. 2020{\natexlab{b}}.
\newblock \href
  {https://proceedings.neurips.cc/paper/2020/hash/4c2e5eaae9152079b9e95845750bb9ab-Abstract.html}
  {What makes for good views for contrastive learning?}
\newblock In \emph{Advances in Neural Information Processing Systems 33: Annual
  Conference on Neural Information Processing Systems 2020, NeurIPS 2020,
  December 6-12, 2020, virtual}.

\bibitem[{Vaswani et~al.(2017)Vaswani, Shazeer, Parmar, Uszkoreit, Jones,
  Gomez, Kaiser, and Polosukhin}]{Vaswani2017}
Ashish Vaswani, Noam Shazeer, Niki Parmar, Jakob Uszkoreit, Llion Jones,
  Aidan~N. Gomez, Lukasz Kaiser, and Illia Polosukhin. 2017.
\newblock \href {https://doi.org/10.1017/CBO9780511809071} {{Attention Is All
  You Need}}.
\newblock \emph{Proceedings of the 31st International Conference on Neural
  Information Processing Systems}, pages 6000--6010.

\bibitem[{Wadden et~al.(2020)Wadden, Lin, Lo, Wang, van Zuylen, Cohan, and
  Hajishirzi}]{Wadden2020}
David Wadden, Shanchuan Lin, Kyle Lo, Lucy~Lu Wang, Madeleine van Zuylen, Arman
  Cohan, and Hannaneh Hajishirzi. 2020.
\newblock \href {https://doi.org/10.18653/v1/2020.emnlp-main.609} {{Fact or
  Fiction: Verifying Scientific Claims}}.
\newblock In \emph{Proceedings of the 2020 Conference on Empirical Methods in
  Natural Language Processing (EMNLP)}, pages 7534--7550, Stroudsburg, PA, USA.
  Association for Computational Linguistics.

\bibitem[{Wang and Isola(2020)}]{PositivesSimilarImportant}
Tongzhou Wang and Phillip Isola. 2020.
\newblock \href {https://proceedings.mlr.press/v119/wang20k.html}
  {Understanding contrastive representation learning through alignment and
  uniformity on the hypersphere}.
\newblock In \emph{Proceedings of the 37th International Conference on Machine
  Learning}, volume 119 of \emph{Proceedings of Machine Learning Research},
  pages 9929--9939. PMLR.

\bibitem[{Wang et~al.(2018)Wang, Ye, and Gupta}]{Wang2018ZeroShotRV}
X.~Wang, Yufei Ye, and Abhinav~Kumar Gupta. 2018.
\newblock Zero-shot recognition via semantic embeddings and knowledge graphs.
\newblock \emph{2018 IEEE/CVF Conference on Computer Vision and Pattern
  Recognition}, pages 6857--6866.

\bibitem[{Wolf et~al.(2020)Wolf, Debut, Sanh, Chaumond, Delangue, Moi, Cistac,
  Rault, Louf, Funtowicz, and Brew}]{Wolf2019}
Thomas Wolf, Lysandre Debut, Victor Sanh, Julien Chaumond, Clement Delangue,
  Anthony Moi, Pierric Cistac, Tim Rault, R{\'{e}}mi Louf, Morgan Funtowicz,
  and Jamie Brew. 2020.
\newblock \href {http://arxiv.org/abs/1910.03771} {{Transformers:
  State-of-the-Art Natural Language Processing}}.
\newblock In \emph{Proceedings of the 2020 Conference on Empirical Methods in
  Natural Language Processing: System Demonstrations}, pages 38--45.

\bibitem[{Wright and Augenstein(2021)}]{wright2021citeworth}
Dustin Wright and Isabelle Augenstein. 2021.
\newblock \href {https://doi.org/10.18653/v1/2021.findings-acl.157}
  {{C}ite{W}orth: Cite-worthiness detection for improved scientific document
  understanding}.
\newblock In \emph{Findings of the Association for Computational Linguistics:
  ACL-IJCNLP 2021}, pages 1796--1807, Online. Association for Computational
  Linguistics.

\bibitem[{Wu et~al.(2017)Wu, Manmatha, Smola, and Krahenbuhl}]{Wu2017}
Chao-yuan Wu, R.~Manmatha, Alexander~J Smola, and Philipp Krahenbuhl. 2017.
\newblock \href {https://doi.org/10.1109/ICCV.2017.309} {{Sampling Matters in
  Deep Embedding Learning}}.
\newblock In \emph{2017 IEEE International Conference on Computer Vision
  (ICCV)}, pages 2859--2867. IEEE.

\bibitem[{Wu et~al.(2019)Wu, Zhang, de~Souza, Fifty, Yu, and
  Weinberger}]{Wu2019}
Felix Wu, Tianyi Zhang, Amauri~Holanda de~Souza, Christopher Fifty, Tao Yu, and
  Kilian~Q. Weinberger. 2019.
\newblock \href {http://arxiv.org/abs/1902.07153} {{Simplifying Graph
  Convolutional Networks}}.
\newblock In \emph{Proceedings of the 36th International Conference on Machine
  Learning}, volume 6861-6871, pages 815--826. PMLR.

\bibitem[{Wu et~al.(2021)Wu, Gao, Zang, Han, Wang, and Hu}]{Wu2021}
Xing Wu, Chaochen Gao, Liangjun Zang, Jizhong Han, Zhongyuan Wang, and Songlin
  Hu. 2021.
\newblock \href {http://arxiv.org/abs/2109.04321} {{Smoothed Contrastive
  Learning for Unsupervised Sentence Embedding}}.
\newblock \emph{arXiv:2109.04321}.

\bibitem[{Wu et~al.(2020)Wu, Wang, Gu, Khabsa, Sun, and Ma}]{Wu2020}
Zhuofeng Wu, Sinong Wang, Jiatao Gu, Madian Khabsa, Fei Sun, and Hao Ma. 2020.
\newblock \href {http://arxiv.org/abs/2012.15466} {{CLEAR: Contrastive Learning
  for Sentence Representation}}.
\newblock \emph{arXiv:2012.15466}.

\bibitem[{Xiong et~al.(2020)Xiong, Xiong, Li, Tang, Liu, Bennett, Ahmed, and
  Overwijk}]{Xiong2020}
Lee Xiong, Chenyan Xiong, Ye~Li, Kwok-Fung Tang, Jialin Liu, Paul Bennett,
  Junaid Ahmed, and Arnold Overwijk. 2020.
\newblock \href {http://arxiv.org/abs/2007.00808} {{Approximate Nearest
  Neighbor Negative Contrastive Learning for Dense Text Retrieval}}.
\newblock In \emph{International Conference on Learning Representations}, pages
  1--16.

\bibitem[{Yang et~al.(2015)Yang, Liu, Zhao, Sun, and Chang}]{Yang2015}
Cheng Yang, Zhiyuan Liu, Deli Zhao, Maosong Sun, and Edward~Y. Chang. 2015.
\newblock \href {https://doi.org/10.5555/2832415.2832542} {Network
  representation learning with rich text information}.
\newblock In \emph{Proceedings of the 24th International Conference on
  Artificial Intelligence}, IJCAI'15, page 2111–2117. AAAI Press.

\end{thebibliography}
